\documentclass[aps,a4paper,pra,twocolumn,nofootinbib,nobalancelastpage,longbibliography,nobibnotes,numbers]{revtex4-1}

\makeatletter
\renewcommand{\fnum@figure}{\textbf{\figurename~\thefigure}}
\makeatother
\renewcommand{\figurename}{Fig.}

\usepackage[verbose=true]{microtype}

\usepackage{amsmath}
\usepackage{amssymb}
\usepackage{txfonts}
\usepackage{mathtools}
\usepackage{upgreek}
\usepackage{bm}
\usepackage{siunitx}

\usepackage[colorlinks,linktocpage,hypertexnames=false,pdfpagelabels,urlcolor=black,citecolor=blue,linkcolor=blue]{hyperref}

\makeatletter
\renewcommand\section{%
  \@startsection{subsection}{1}{0pt}%
  {-\baselineskip}{.2\baselineskip}%
  {\normalfont\normalsize\bfseries\raggedright\color{blue}}}
\renewcommand\subsection{%
  \@startsection{subsubsection}{2}{0pt}%
  {-\baselineskip}{.1\baselineskip}%
  {\normalfont\normalsize\bfseries\raggedright}}

\renewcommand\paragraph{%
  \@startsection{paragraph}{4}{0pt}%
  {3.25ex\@plus 1ex\@minus .2ex}{-1em}%
  {\normalfont\normalsize\bfseries}}
\makeatother

\begin{document}

\title{Fractal basin geometry and the limits to predictability and reproducibility in deep learning}

\author{Andrew Ly}
\affiliation{School of Physics, University of Sydney, Sydney, NSW, Australia}
\author{Pulin Gong}
\email{pulin.gong@sydney.edu.au}
\affiliation{School of Physics, University of Sydney, Sydney, NSW, Australia}

\begin{abstract}
  {Fundamental limits to predictability are central to our understanding of many physical and computational systems. In deep learning, training outcomes are often difficult to predict and reproduce even when extrinsic sources of randomness are controlled, yet the underlying mechanism remains unclear. Here we propose riddled basin geometry as a unifying dynamical systems mechanism: every initialization leading to one solution lies arbitrarily close to initializations leading to different solutions, forming a fat fractal with structure at arbitrarily fine scales. Using a minimal neural network model, we analytically establish the classical route to riddling and derive a near-zero uncertainty exponent: even sharp increases in initialization precision yield only marginal gains in outcome predictability. In deep networks, intertwined training destinations and near-zero parameter-space exponents provide robust signatures consistent with riddling across the tested architectures, optimizers and training protocols, offering a common dynamical account of otherwise disparate empirical observations. We further introduce functional uncertainty exponents, extending these diagnostics of scale-insensitive irreproducibility from parameter space to the learned functions themselves. Together, our results support riddled basin geometry as a candidate organizing principle for irreproducibility in deep learning, with practical implications for optimization and the reliable deployment of artificial intelligence.
  }
\end{abstract}

\maketitle

Fundamental limits to predictability are central to physics and computation, from the probabilistic outcomes of quantum measurement \cite{sakurai2020modern}, to the finite forecast horizons of chaotic systems \cite{ott2002chaos}, to the undecidability of the halting problem in universal computation \cite{sipser1996introduction}. An equally important question confronts modern artificial intelligence (AI): what intrinsic limits constrain the predictability and hence reproducibility of training outcomes, even when all extrinsic randomness is controlled? Outcomes that cannot be predicted with certainty are seldom reproduced reliably \cite{sommerer1993physical}; unpredictability and irreproducibility are therefore tightly linked. Addressing these intrinsic limits is essential for both understanding AI systems and their practical deployment. 

A large body of work attributes the unpredictability of training outcomes to extrinsic stochasticity \cite{gundersen2022sources, bhojanapalli2021reproducibility, ahn2022reproducibility, zhuang2022randomness, dodge2020fine, jiang2021churn, yuan2025give}, including random initialization, mini-batch ordering, data augmentation, and numerical errors introduced by the computing platform (e.g., GPU). These studies primarily measure the impact of such noise sources, showing that irreproducibility is challenging to alleviate across diverse architectures and tasks. Surprisingly, however, empirical evidence indicates that even when extrinsic non-determinism is eliminated, changing a parameter by as little as one bit can produce variability comparable to that arising from multiple noise sources combined \cite{summers2021nondeterminism}. 
The mechanism by which such vanishing perturbations yield qualitatively different training outcomes remains unclear, and the deeper question of what intrinsic limits exist on predicting training outcomes remains unaddressed.

\begin{figure*}[t]
    \centering
    \includegraphics{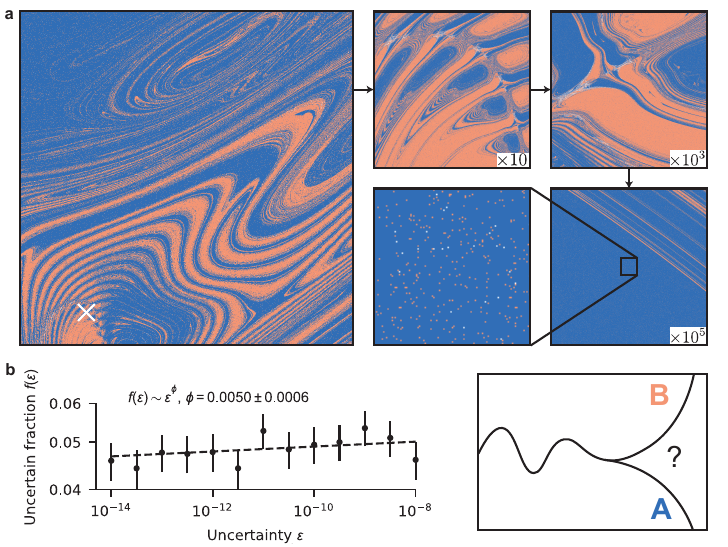}
    \caption{\textbf{Schematic of riddled basins and outcome unpredictability.} \textbf{a,} The basin of attractor $A$ (blue) is riddled with that of attractor $B$ (orange). Arrows indicate successive magnifications centered on the white cross; the final panel zooms in on the boxed region, revealing interleaved fractal structure at arbitrarily fine scales. \textbf{b,} The fractal structure is quantified by $f(\varepsilon)$, the probability that a random perturbation of magnitude $\varepsilon$ changes the basin. Dots and error bars represent the mean and $95\%$ confidence intervals from $n=10^4$ $\varepsilon$-close pairs (i.e., pairs of initializations separated by at most $\varepsilon$ in each dimension). A near-zero uncertainty exponent $\phi$, defined by $f(\varepsilon) \sim \varepsilon^\phi$, indicates that increasing the precision of the initialization yields only marginal gains in predictability; the qualitative fate of a given initialization ($A$ or $B$?) remains effectively unpredictable, thus undermining reproducibility.}
    \label{fig:1}
\end{figure*}

Here we propose riddled basin geometry as a unifying dynamical-systems mechanism for this unpredictability: the basins of attraction that govern training outcomes in parameter space are intertwined, such that any initialization leading to one solution lies arbitrarily close to initializations leading to a different solution  \cite{aguirre2009fractal, alexander1992riddled} (Fig.~\ref{fig:1}), reflecting a ``fat'' fractal organization.
This differs from previously reported trainability boundaries involving ``skinny'' fractals of zero Lebesgue measure \cite{sohl2024boundary}, where predictability is recoverable by sufficiently decreasing the uncertainty, $\varepsilon$, of the initial conditions. 
In contrast, fat fractality makes a riddled basin inseparable from its boundary, so that no increase in precision restores reliable prediction of the training outcome.
Riddling thus entails a form of unpredictability that goes beyond the chaotic behavior emphasized in recent studies \cite{kong2020stochasticity, herrmann2022chaotic, ly2025optimization}: chaos limits trajectory-level predictability, with small errors in the initial conditions growing over time, whereas riddling yields qualitative, outcome-level unpredictability that persists under arbitrarily small uncertainty \cite{sommerer1993physical, sommerer1996intermingled}.

We first illustrate how riddling can emerge in learning dynamics, using a minimal model that analytically links properties widespread in deep learning, including symmetries \cite{ziyin2025parameter} and convergence towards symmetry-induced invariant subspaces \cite{chen2023stochastic}. To quantify fine-scale fractal structure, we adapt uncertainty exponents from dynamical systems theory \cite{grebogi1983final, mcdonald1985fractal}. We find near-zero exponents consistent with riddled basin geometry, indicating that reproducibility does not materially improve with increasing initialization precision.
We then generalize to the parameter space of deep networks, showing that the same riddling-like signatures arise in realistic settings. Because practical applications prioritize a model's input-output behavior over its specific parameterization \cite{jiang2021churn, zhuang2022randomness}, we further introduce functional analogues of the uncertainty exponent, based on standard measures of functional similarity \cite{klabunde2025similarity}, and demonstrate that scale-insensitive irreproducibility extends to function space, as signified by near-zero functional exponents. 
These findings are robust across tested configurations, including different architectures, optimizers, regularization methods and activation functions. Finally, we show that the riddling framework rationalizes disparate training phenomena, including irreducible model variability \cite{bhojanapalli2021reproducibility, zhuang2022randomness, dodge2020fine, yuan2025give, summers2021nondeterminism} and performance-reproducibility trade-offs \cite{cohen2021gradient, herrmann2022chaotic}. Together, our results establish a robust scale-insensitive limit to reproducibility in deep learning and support riddled basin geometry as a candidate organizing principle underlying it.

\section*{Neural network training}  

A neural network, $f_{\bm{\uptheta}}: X \rightarrow Y$, is parameterized by $\bm \uptheta \in \mathbb{R}^d$ representing its vectorized state in parameter space. Training involves minimizing the error in approximating a dataset $S = \{(\mathbf{x}_i,\mathbf{y}_i)\}_{i=1}^N \subset X \times Y$, quantified by the loss:
\begin{equation}
    L(\bm{\uptheta}) = \frac{1}{N} \sum_{i=1}^N l(f_{\bm{\uptheta}}(\mathbf{x}_i), \mathbf{y}_i), 
\end{equation}
where $l$ is the single-sample loss function. Standard training algorithms minimize the loss by iteratively updating $\bm\uptheta$. For example, the stochastic gradient descent (SGD) algorithm can be expressed as the discrete-time dynamical system, $\Phi$, given by: 
\begin{equation}
    \Phi(\bm{\uptheta}_t) = \bm{\uptheta}_{t+1} = \bm{\uptheta}_t - \frac{\eta}{b} \sum_{i\in B_t}\nabla_{\bm\uptheta} l(f_{\bm{\uptheta}}(\mathbf{x}_i), \mathbf{y}_i),
    \label{eq:sgd}
\end{equation}
where $t$ denotes the iteration, $\eta$ is the learning rate and $B_t \subset \{1,\dots,N\}$ are mini-batches of size $|B_t| = b$ for all $t$. The learning rate $\eta$ and mini-batch size $b$ are hyperparameters that govern the training dynamics.

Recent studies have shown that training dynamics are constrained by symmetries \cite{ziyin2025parameter, ziyin2024parameter, chen2023stochastic, simsek2021geometry}---parameter transformations that preserve the network function. These symmetries induce invariant subspaces: trajectories starting within them remain there indefinitely. Randomly initialized networks have been observed to converge to these subspaces \cite{chen2023stochastic}, implying the presence of attractors embedded within them. Because symmetries are abundant in deep networks \cite{ziyin2025parameter}, many such attractors can coexist, and numerical evidence indicates that they are chaotic \cite{ly2025optimization, kong2020stochasticity}. Analytic results show that their stability can be weakened by large learning rates (e.g., the edge of stability phenomenon \cite{cohen2021gradient}) or by negative curvatures in the loss landscape \cite{herrmann2022chaotic}. These three elements---symmetry-induced invariant subspaces, chaotic attractors and weakened stability---are not isolated features but interlinked. Together, these ingredients define a theoretically established route to riddling \cite{alexander1992riddled}, and given their ubiquity in deep learning, one may expect riddling could be common. 

\begin{figure*}[t]
    \centering
    \includegraphics{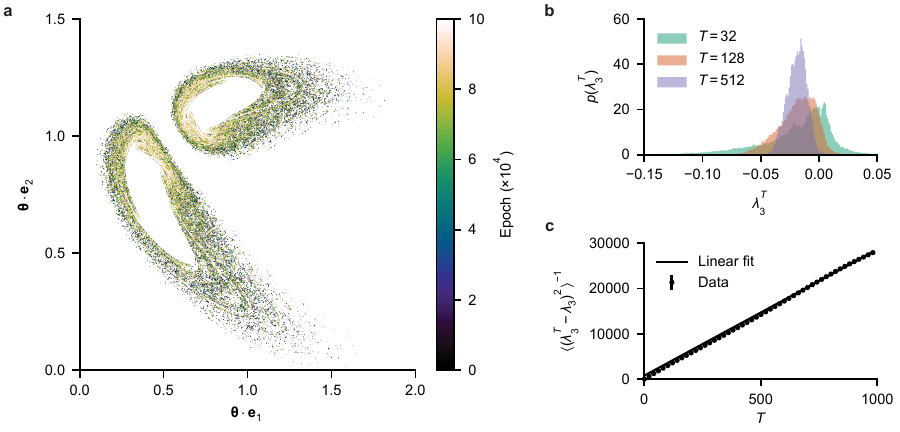}
    \caption{\textbf{Chaotic attractor in the training of the minimal model.} \textbf{a,} A chaotic attractor within the permutation-invariant plane $\mathcal{P}_{+}$ is traced by the training trajectory from a random initialization $\bm\uptheta_0 \in \mathcal{P}_{+}$ (see ``Methods'' for details). Each point represents the coordinates of an iterate with respect to the basis of $\mathcal{P}_{+}$, comprising $\mathbf{e}_1 = (1,1,0,0)/\sqrt{2}$ and $\mathbf{e}_2 = (0,0,1,1)/\sqrt{2}$; color encodes epoch. \textbf{b,} Distributions of finite time-$T$ transverse Lyapunov exponents, $\lambda_3^T$ for $T = 32, 128, 512$, show non-zero fractions of positive values: $28.9\%$, $9.8\%$ and $0.7\%$, respectively. \textbf{c,} The inverse mean squared fluctuations of finite-time exponents, $\langle (\lambda_3^T - \lambda_3)^2\rangle^{-1}$, grows linearly with $T$ for large $T$. Error bars, which denote 95\% confidence intervals, are smaller than the points.}
    \label{fig:2}
\end{figure*}

\section*{Riddled basins in neural network training}

In this section, we establish that neural network training can satisfy the theoretical conditions sufficient for riddling \cite{alexander1992riddled}: a chaotic attractor exists within an invariant subspace, which contains a zero-measure set of transversely unstable periodic points, together with at least one competing attractor elsewhere (see ``Methods'' for a dynamical systems primer). 
Because direct stability analysis of high-dimensional attractors in deep network training is analytically and computationally intractable, we adopt a minimal modeling approach. 
This approach, widely used in theoretical studies of deep learning \cite{chen2023stochastic, ly2025optimization, bahri2020statistical, baldassi2021unveiling}, provides tractability while yielding key conceptual insights. 

We choose our minimal model to be a two-layer network since it is the smallest architecture that admits a non-trivial symmetry: 
Specifically, it is an instance of the two-layer networks studied in a classic theoretical work \cite{mei2018mean}; we write the network function in a form that parallels their formulation: 
\begin{equation}
    f_{\bm\uptheta}(x) = \alpha^{(2)} \mathbf{w}^{(2)} \sigma(\alpha^{(1)} \mathbf{w}^{(1)} x) = \sum_{i=1}^2 \alpha^{(2)} w^{(2)}_{i} \sigma(\alpha^{(1)}w^{(1)}_i x), \label{eq:2nn}
\end{equation}
where $\bm{\uptheta} = (\mathbf{w}^{(1)\top}, \mathbf{w}^{(2)})^{\top} = (w^{(1)}_{1}, w^{(1)}_{2}, w^{(2)}_{1}, w^{(2)}_{2}) \in \mathbb{R}^4$ is the vector of network weights with the subscript and superscript indexing the neurons and layers, respectively. Each neuron's weights are denoted as $\mathbf{w}_i = (w^{(1)}_i, w^{(2)}_i)$. We use the hyperbolic tangent activation, $\sigma = \tanh$, which under the mean field parameterization yields scaling factors \cite{mei2018mean}: $\alpha^{(1)} = \sqrt{2}$ and $\alpha^{(2)} = 1/2$. We train the network to perform regression on $S = \{(x_i,y_i)\}_{i=1}^{8}$, where $x_i, y_i \sim \mathcal{N}(0,1)$. Specifically, we apply deterministic gradient descent (GD) to minimize the mean squared error, $L(\bm\uptheta) = \frac{1}{8} \sum_{i=1}^{8}(f_{\bm\uptheta}(x_i) - y_i)^2$. 

\textbf{Symmetry-induced invariant subspaces.} 
We begin by analytically identifying invariant subspaces and training destinations in our minimal model. In total, the dynamical system governing training contains four symmetry-induced invariant subspaces (see ``Methods''). First, permutation symmetry generates a two-dimensional invariant plane, $\mathcal{P}_{+} \coloneqq \{\bm{\uptheta} \in \mathbb{R}^4 \mid \mathbf{w}_1 = \mathbf{w}_2\}$, where the two hidden neurons are identical. Second, the odd symmetry of the activation function (i.e., $\tanh(-x) = -\tanh(x)$) yields an additional permutation-invariant plane, $\mathcal{P}_{-} \coloneqq \{\bm{\uptheta} \in \mathbb{R}^4 \mid \mathbf{w}_1 = - \mathbf{w}_2\}$, representing a sign difference. Finally, the origin-passing activation (i.e., $\tanh(0) = 0$) induces two parity-invariant planes \cite{chen2023stochastic}, $\mathcal{P}_{0}^{i} \coloneqq \{\bm\uptheta \in \mathbb{R}^4 \mid \mathbf{w}_i = 0\}$ for $i = 1,2$, corresponding to vanishing neurons. We find that the predominant competing destinations among random initializations are the permutation-invariant planes, $\mathcal{P}_{\pm}$ (Extended Data Fig.~1). At large learning rates, another possible outcome is divergence, $\Vert \bm\uptheta_t \Vert \rightarrow \infty$, which is treated as an attractor at infinity \cite{sommerer1993physical}.

\begin{figure*}[t]
    \centering
    \includegraphics{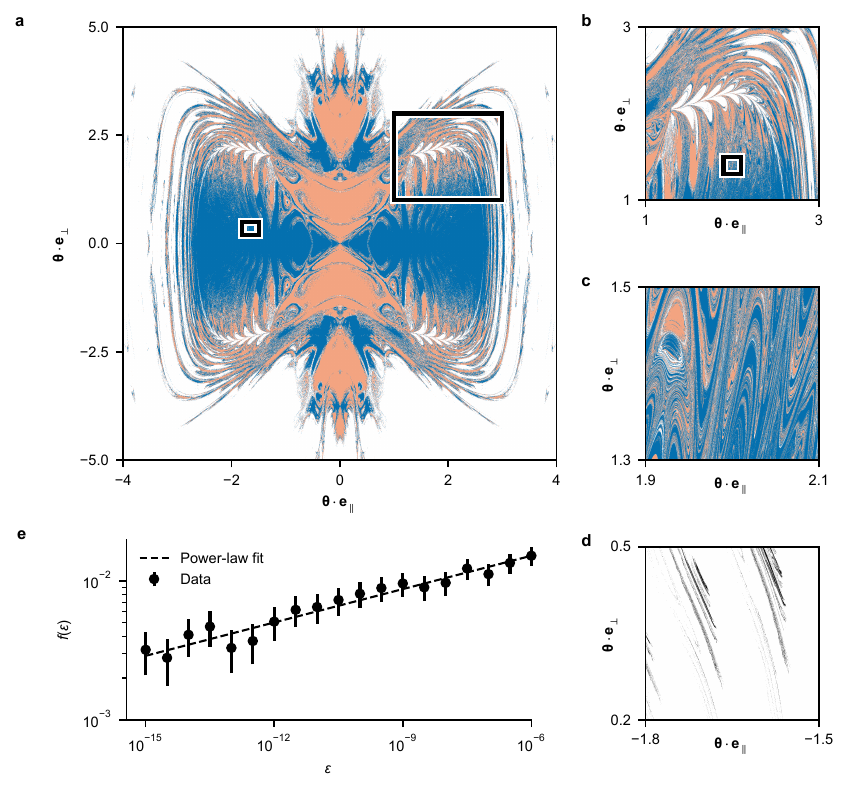}
    \caption{\textbf{Riddling in the minimal model.} \textbf{a,} Destination map for a $2047 \times 2047$ uniform grid of initializations on the plane spanned by two random directions $\mathbf{e}_{\parallel}$ and $\mathbf{e}_{\perp}$, which are longitudinal and transverse to the $\mathcal{P}_{+}$ permutation-invariant subspace, respectively. Initializations converging to $\mathcal{P}_{+}$, $\mathcal{P}_{-}$ and infinity are colored blue, orange and white, respectively. The resulting basins of attraction exhibit a striking butterfly-like pattern. \textbf{b,} Magnification of the right inset in (\textbf{a}) on a $1024 \times 1024$ grid. \textbf{c,} Magnification of the inset in (\textbf{b}) on a $1024 \times 1024$ grid. \textbf{d,} Magnification of the left inset in (\textbf{a}) on a $1024 \times 1024$ grid; for visibility of fine-scale structure, blue and orange are replaced with white and black, respectively. \textbf{e,} The uncertainty fraction $f(\varepsilon)$ exhibits small-$\varepsilon$ scaling $f(\varepsilon) \sim \varepsilon^{\phi_p}$ with uncertainty exponent $\phi_p = 0.08003\pm 0.00007$. Dots and error bars show the mean and $95\%$ confidence intervals from $n=10^4$ $\varepsilon$-close pairs.}
    \label{fig:3}
\end{figure*}

\textbf{Chaotic attractor.} At a learning rate of $\eta = 2.7$, the $\mathcal{P}_{+}$ plane exhibits an attractor $A$ (Fig.~\ref{fig:2}\hyperref[fig:2]{a}). We assess the stability of $A$ in terms of the longitudinal and transverse Lyapunov exponents of equation~(\ref{eq:sgd}) (see ``Methods''). Intuitively, these exponents quantify the exponential rates of divergence on $A$ in directions along and away from $\mathcal{P}_+$, respectively. The longitudinal exponents converge to $\lambda_1 = 0.1192$ and $\lambda_2 = -0.1039$. The positive maximal exponent confirms that $A$ is a chaotic attractor \cite{sommerer1993physical}. Conversely, the transverse exponents converge to $\lambda_3 = -0.0190$ and $\lambda_4 = -0.1811$. Since all transverse exponents are negative, $A$ is a Milnor attractor in the full parameter space \cite{milnor1985concept, alexander1992riddled}. That is, its basin of attraction, $\beta(A)$, has non-zero four-dimensional Lebesgue measure.

\textbf{Weakened transverse stability.} We next reveal a zero measure set of transversely unstable periodic points embedded in the chaotic attractor $A$. The explicit determination of such points is possible for only a few dynamical systems \cite{aguirre2009fractal}. However, since they cause unstable dimension variability, it is possible to infer their presence by observing positive fluctuations of the finite-time transverse Lyapunov exponents \cite{aguirre2009fractal}. We calculate the time-$T$ value, $\lambda_3^T$, by partitioning a long trajectory into segments of length $T$ (see ``Methods''), confirming a non-zero fraction of positive fluctuations for various $T$ (Fig.~\ref{fig:2}\hyperref[fig:2]{b}). Moreover, the fluctuations of finite-time exponents exhibit the scaling: 
\begin{equation}
    \langle (\lambda_3^T - \lambda_3)^2 \rangle = \frac{2D}{T} \quad \text{for large $T$,}
\end{equation}
where $D = 0.0179 \pm 0.0002$ is the diffusion coefficient, estimated by weighted least squares regression for $T \in [600,1000]$ (Fig.~\ref{fig:2}\hyperref[fig:2]{c}). This diffusive scaling is a hallmark of systems that exhibit riddling \cite{ott1994transition}. Altogether, the results indicate a chaotic attractor within the $\mathcal{P}_+$ invariant subspace, containing transversely unstable periodic points, that competes with attractors in $\mathcal{P}_-$ and infinity. The minimal model satisfies all theoretical conditions sufficient for the emergence of riddling. 

\textbf{The emergence of riddling.} With these conditions satisfied, dynamical systems theory predicts that the basin $\beta(A)$ is riddled; see ``Methods'' and Extended Data Fig.~2 for details on the route to riddling. To confirm this prediction in the minimal model, we determine the training outcome across a high-resolution grid of initializations. The grid is taken in a plane spanned by random orthonormal vectors, $\mathbf{e}_{\parallel}$ and $\mathbf{e}_{\perp}$, longitudinal and transverse to $\mathcal{P}_{+}$, respectively. Figure~\ref{fig:3}\hyperref[fig:3]{a} shows the eventual destination of each initialization, revealing an intricate butterfly-like pattern. Convergence to $\mathcal{P}_{+}$ is interwoven with other outcomes, even in regions that appear uniform at a coarse resolution (Fig.~\ref{fig:3}\hyperref[fig:3]{a-d}). Increasing magnification uncovers complexity across many scales, demonstrating the indefinite scaling of a riddled basin's fat-fractal geometry \cite{ott1993scaling}. The basin of $\mathcal{P}_{-}$ exhibits a similar structure, with points leading to $\mathcal{P}_{+}$ densely embedded throughout. Being mutually riddled, the two basins are intermingled \cite{sommerer1996intermingled}.

\textbf{A limit to reproducibility.} We next quantify the fine-scale structure of riddling and show its profound consequences for the predictability of the training outcome. We measure the fraction of $\varepsilon$-uncertain initializations, $f(\varepsilon)$, defined as the probability that an uncertainty $\varepsilon$ in the initialization of two otherwise identical training runs leads to differing outcomes in parameter space \cite{grebogi1983final, mcdonald1985fractal}. For fractal basin boundaries, this fraction scales as a power law at small $\varepsilon$ \cite{grebogi1983final, aguirre2009fractal}:
\begin{equation}
    f(\varepsilon) \propto \varepsilon^{\phi_p},
\end{equation}
where $\phi_p$ is the uncertainty exponent. For initializations at a fixed distance away from the $\mathcal{P}_{\pm}$ planes (see ``Methods''), we find that $\phi_p = 0.08003\pm 0.00007$ by weighted least squares regression (Figure~\ref{fig:3}\hyperref[fig:3]{e}). This near-zero value implies a reduction of $\varepsilon$ by almost $4$ orders of magnitude is needed to merely halve $f(\varepsilon)$. Thus, arbitrarily small perturbations to training can alter the training outcome with non-vanishing probability; because this sensitivity persists at all scales, increasing initialization precision---with all other factors controlled---does not materially improve reproducibility.

Although some sources of stochasticity are manageable by fixing seeds (e.g., initialization, mini-batch ordering, data augmentation, etc.), there is often unavoidable non-deterministic perturbations, including those introduced by the computing platform (e.g., GPU). In such cases, even seemingly identical training runs can produce considerable difference between network predictions \cite{bhojanapalli2021reproducibility, summers2021nondeterminism, milani2016launch, jiang2021churn, zhuang2022randomness}. 
Our results establish riddling as a potential dynamical mechanism for explaining these reproducibility limits in deep learning.

\section*{Deep neural network training}
In this section, we examine riddling-like irreproducibility in deep networks.
First, we identify signatures of riddling in parameter space, including near-zero uncertainty exponents, $\phi_p \approx 0$. 
Second, we introduce functional uncertainty exponents, $\phi_f$, as a diagnostic of reproducibility limits in function space. 
Third, we establish riddling analysis as a practical diagnostic framework for irreproducibility in broader settings.

\textbf{Parameter-space irreproducibility.} In total, we train approximately \num[group-separator={,}]{500000} networks spanning a wide range of training configurations, including different architectures such as VGGNet \cite{simonyan2014very}, ResNet \cite{he2016deep} and BERT \cite{devlin2019bert}; datasets and modalities, such as MNIST and CIFAR-10 image classification and IMDB sentiment classification; optimization algorithms, such as SGD, GD and AdamW \cite{loshchilov2017decoupled}; regularization methods, such as L1 and L2 regularization, batch and layer normalization, and dropout; and activation functions, such as $\tanh$, ReLU and GELU. 
Here, we focus on a representative example: the training of the VGG-12 network with $\tanh$ activations on the MNIST dataset corrupted with $50\%$ label noise \cite{frenay2013classification}, using SGD with momentum and weight decay to minimize the cross-entropy loss (see ``Methods''). 
This setting is computationally efficient, allowing high-resolution sampling and robust statistics. Importantly, however, the same qualitative results are observed across all configurations, including those without label noise (Supplementary Sec.~8). 

\begin{figure*}[t]
    \centering
    \includegraphics{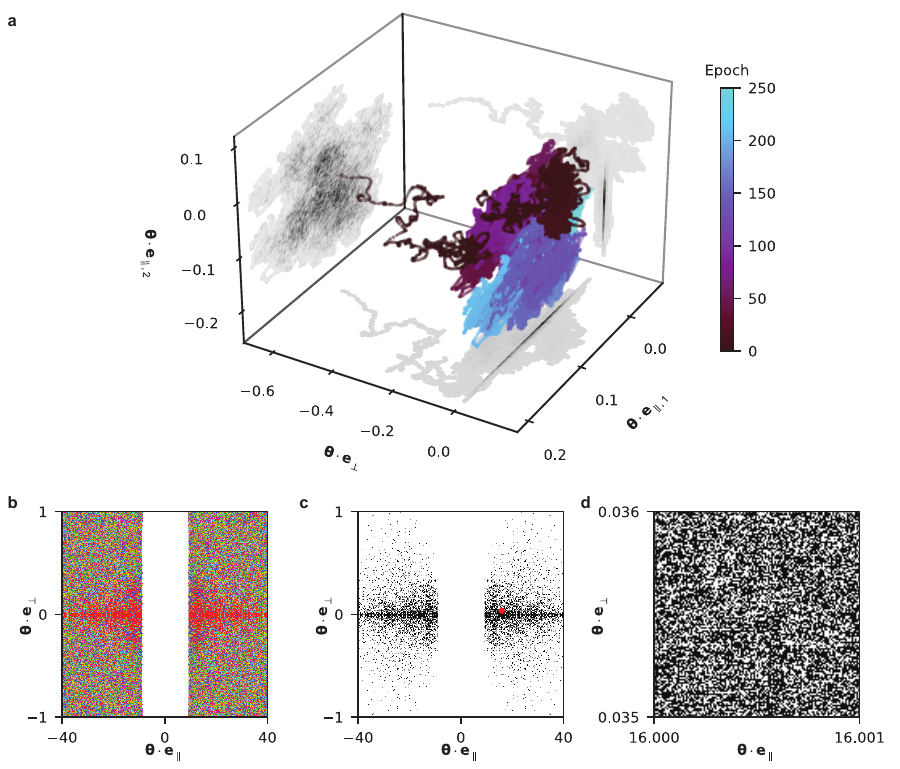}
    \caption{\textbf{Riddling-like dynamics of deep neural network training.} \textbf{a,} A randomly initialized VGG-12 network is attracted to a parity-invariant subspace during training. The vectorized network weights $\bm{\uptheta}$ are projected onto three random dimensions: $\mathbf{e}_{\parallel,1}$, $\mathbf{e}_{\parallel,2}$ (longitudinal to the invariant subspace) and $\mathbf{e}_{\perp}$ (transverse). The shadow on each pane is a two-dimensional histogram, where darker shades indicate higher frequency. \textbf{b,} Training destination map for a $255 \times 255$ uniform grid of initializations on the plane spanned by $\mathbf{e}_{\parallel} = \mathbf{e}_{\parallel,1}$ and $\mathbf{e}_{\perp}$. Each color denotes a unique parity-invariant subspace; in total, there are 1772 different destinations. White points approach the origin. \textbf{c,} Same as (\textbf{b}), except initializations converging to the invariant subspace at $\bm{\uptheta} \cdot \mathbf{e}_\perp = 0$ are colored black. All other destinations are white. \textbf{d,} Magnification around the red dot in (\textbf{c}) on a $128 \times 128$ grid.}
    \label{fig:4}
\end{figure*}

In this representative example, training trajectories converge to parity-invariant subspaces wherein multiple neurons in the last hidden layer vanish (Fig.~\ref{fig:4}\hyperref[fig:4]{a}); see ``Methods'' for proof of invariance. To visualize the competition between different outcomes, we train a grid of networks. The destination map (Fig.~\ref{fig:4}\hyperref[fig:4]{b}) shows extensive intermixing of diverse outcomes, including 1772 different parity-invariant subspaces, identified by which neurons vanish. Beneath the heterogeneity is a diffuse structure representing the basin of attraction to the invariant subspace at $\bm{\uptheta} \cdot \mathbf{e}_{\perp} = 0$ (Fig.~\ref{fig:4}\hyperref[fig:4]{c}). Qualitatively, this basin appears riddled; it is perforated with initializations leading elsewhere. A further magnification (Fig.~\ref{fig:4}\hyperref[fig:4]{d}) uncovers intermixing at smaller scales, consistent with riddling-like, fat fractality.

\begin{figure*}[t]
    \centering
    \includegraphics{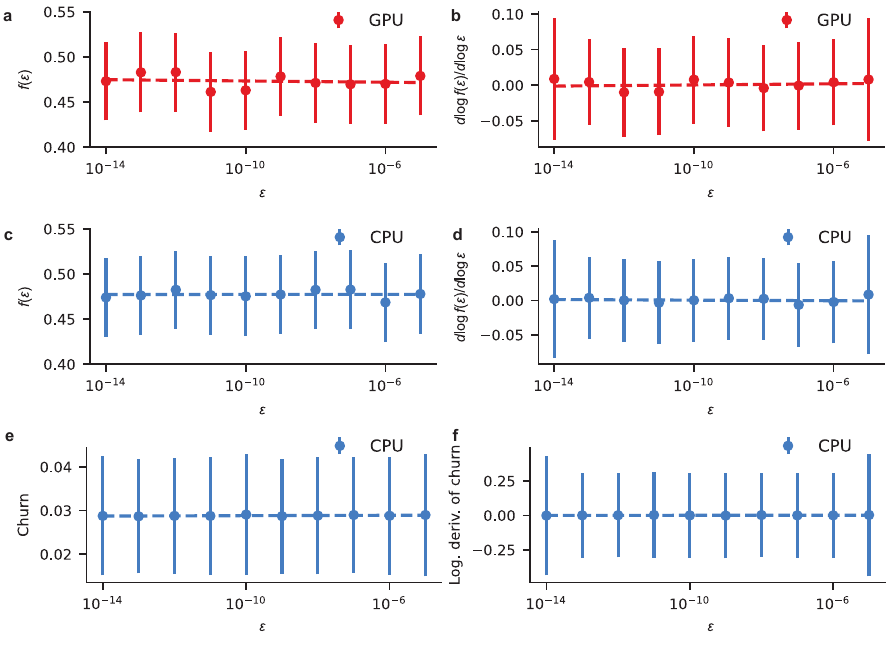}
    \caption{\textbf{Parameter- and function-space uncertainty exponents.} \textbf{a,} The uncertainty fraction $f(\varepsilon)$ for initializations within a $\varepsilon$-hypercube centered at the middle of Fig.~\ref{fig:4}\hyperref[fig:4]{d}, $(\bm{\uptheta}\cdot \mathbf{e}_{\parallel}, \bm{\uptheta}\cdot \mathbf{e}_{\perp}) = (16.0005, 0.0355)$. Networks are trained on GPU, which is faster than CPU but non-deterministic. Dots and error bars show the mean and $95\%$ confidence intervals from $10^4$ bootstrap resamples, each of size $n=500$ $\varepsilon$-close pairs. A power-law fit $f(\varepsilon) \propto \varepsilon^{\phi_p}$ (dashed line) yields the uncertainty exponent $\phi_p = 0.000 \pm 0.002$. \textbf{b,} The logarithmic derivative of (\textbf{a}). \textbf{c,} Same as (\textbf{a}), except with CPU training to ensure determinism. The uncertainty exponent is also $\phi_p = 0.000 \pm 0.002$. \textbf{d,} The logarithmic derivative of (\textbf{c}). \textbf{e,} Dots and error bars denote the mean churn and $95\%$ confidence interval across $n=250$ CPU-trained $\varepsilon$-close pairs. The functional uncertainty exponent using churn is $\phi_f = 0.00 \pm 0.01$ and the mean churn plateaus at approximately $2.9\%$. \textbf{f,} The logarithmic derivative of (\textbf{e}).}
    \label{fig:5}
\end{figure*}

We next quantify the irreproducibility of parameter-space outcomes. Specifically, we calculate the fraction $f(\varepsilon)$ of $\varepsilon$-uncertain initializations near $(\bm{\uptheta} \cdot \mathbf{e}_{\parallel}, \bm{\uptheta} \cdot \mathbf{e}_{\perp}) = (16.0005, 0.0355)$, which is the center of Fig.~\ref{fig:5}\hyperref[fig:5]{d}, using both GPU (Fig.~\ref{fig:5}\hyperref[fig:5]{a}) and CPU (Fig.~\ref{fig:5}\hyperref[fig:5]{c}). For both platforms, the best power-law fit by weighted least squares regression across 10 orders of magnitude exhibits an exponent of $\phi_p = 0.000 \pm 0.002$. Local estimates obtained from the logarithmic derivative confirm that $\phi_p$ remains near zero even at the smallest accessible scales (Figs.~\ref{fig:5}\hyperref[fig:5]{b, d}). The two platforms play complementary roles: CPU enables deterministic computation, ruling out the possibility that the observed irreproducibility is an artifact of accumulated random error, whereas GPU reflects a realistic deep-learning setting, in which acceleration comes at the cost of non-determinism \cite{krizhevsky2012imagenet}. Together, these results show a parameter-space reproducibility limit consistent with riddling.

\textbf{Function-space irreproducibility.} In practice, it is the reproducibility of the learned input-output mapping (i.e., the function) of a network that matters \cite{summers2021nondeterminism, bhojanapalli2021reproducibility}. However, parameter-space irreproducibility does not imply function-space irreproducibility due to the presence of symmetries that enable different parameter configurations to represent the same function \cite{hecht1990algebraic, chen2023stochastic}, and for specific networks (e.g., linear), imply convergence to initialization-independent solutions \cite{ziyin2024parameter}. This raises the question: does functional reproducibility exhibit analogous limits?

To address this question, we introduce the functional uncertainty exponent, $\phi_f$, based on the power-law scaling of a functional similarity measure: $m_{\varepsilon} \propto \varepsilon^{\phi_f}$, where $m_\varepsilon$ is the average similarity of two networks trained from $\varepsilon$-close initializations. 
Unlike the uncertainty exponent $\phi_p$, the functional analogue $\phi_f$ does not quantify basin geometry. It diagnoses whether predictive differences diminish as initialization precision increases.
Here we use churn, defined as the fraction of inputs from dataset $X$ on which two trained networks make different predictions:
\begin{equation}
    m_{\varepsilon}(\mathbf{O}, \mathbf{O}') = \mathbb{E}_X\left[\text{arg\,max}_j O_{i,j} \neq \text{arg\,max}_j O'_{i,j} \right],
\end{equation}
where $\mathbf{O}, \mathbf{O}' \in \mathbb{R}^{N \times C}$ are output matrices of two networks, with entries $O_{i,j}$ and $O'_{i,j}$ denoting the $j$-th class logit for the $i$-th data input. 
Strikingly, increasing initialization precision does not improve functional reproducibility: the churn plateaus at approximately 2.9\% and the functional exponent is indistinguishable from zero, $\phi_f = 0.00 \pm 0.01$ (Figs.~\ref{fig:5}\hyperref[fig:5]{e, f}). We find consistent results across 32 functional measures from the literature \cite{klabunde2025similarity}, including Cohen's kappa \cite{cohen1960coefficient}, centered kernel alignment \cite{kornblith2019similarity} and representational similarity analysis \cite{kriegeskorte2008representational} (Extended Data Fig.~3 and Supplementary Sec.~5). 
Thus, scale-insensitive irreproducibility is not confined to parameter-space outcomes; functional outcomes exhibit an analogous reproducibility limit.

With sufficient computational resources, our riddling analysis can be generalized to state-of-the-art networks. If network symmetries and their corresponding invariant subspaces are known, then it is possible to compute the uncertainty exponent in parameter space to ascertain evidence of riddling. In practice, this can be challenging for two reasons: First, identifying the complete set of symmetry-induced invariant subspaces is an open problem for many architectures \cite{zhao2025understanding}. Second, evaluating the training destination can be computationally intensive. For example, one must compare every possible intra-layer pair of neurons to identify the permutation-invariant subspace to which a network converges. 

We thus propose functional analysis as a practical and architecture-agnostic diagnostic of riddling-like behavior. A training configuration is considered to be in a riddling-like regime when two conditions are satisfied: First, the functional uncertainty exponent is near zero. Second, nearby initializations yield substantial functional differences. Specifically, within the regime exhibiting parameter-space irreproducibility (i.e., $\phi_p \approx 0$), we find that the magnitude of functional difference correlates strongly with generalization, irrespective of configuration details; outside it, functional differences are negligible and carry no meaningful correlation with generalization (Extended Data Fig.~4 and Supplementary Sec.~6). Applying this diagnostic systematically, we find that the signatures of riddling-like behavior are robust to design choices spanning architecture, dataset, training algorithm, regularization and activation function (Extended Data Figs.~5 and 6; Supplementary Sec.~7 and 8), as well as methodological variables, including specific parameter-space slices, reference points, training horizons, thresholds and fitting ranges (Supplementary Sec.~9). 

\section*{Practical implications of the riddling framework} 
We next elucidate the practical implications of the riddling framework. In particular, it predicts a form of unpredictability stronger than chaos and provides a unified dynamical-systems framework for understanding seemingly disparate phenomena in deep learning. 

\textbf{Unpredictability beyond chaos.} 
The riddling framework predicts a form of unpredictability that is radically stronger than chaos. Chaotic systems are quantitatively unpredictable: small errors grow exponentially, limiting the ability to forecast detailed trajectories. Yet they often remain qualitatively predictable since the long-term evolution stays near a single attractor. Qualitative predictability breaks down when multiple attractors compete and the boundary between their basins is fractal. In typical systems with fractal boundaries \cite{mcdonald1985fractal}, almost every initialization has a neighborhood entirely contained in its basin. In such cases, the eventual destination can, in principle, be determined exactly given perfect knowledge of its initial state; that is, $\lim_{\varepsilon \rightarrow 0^+} f(\varepsilon) = 0$, where $f(\varepsilon)$ denotes the probability of misclassification due to uncertainty $\varepsilon$ in the initial state. Riddling represents a drastic departure from this. In a riddled basin, every initialization has a neighborhood containing a positive-measure intersection of its complement \cite{alexander1992riddled}. This geometry obstructs the prediction of long-term behavior from finite-precision information. Thus, it undermines practical applications, including safety-critical ones like medical diagnostics \cite{jiang2021churn, zhuang2022randomness}, which require that small changes to training do not drastically change the learned model.

\textbf{Irreducible model variability.} The riddling framework also predicts that outcome variability is insensitive to the scale of differences between training runs. In practice, this is true only up to machine precision, since network parameterizations separated by less than that are represented identically. 
We confirm this prediction by analyzing the ensemble of possible least-significant-bit flips to the reference network in Fig.~\ref{fig:5}, representing perturbations by the smallest possible amounts within machine precision.
Indeed, the fraction of $\varepsilon$-uncertain pairs, where $\varepsilon$ represents the least significant bit, agrees with the near-constant value in Fig.~\ref{fig:5}: $f(\varepsilon) = 0.477$. Simultaneously, we observe that extrinsic perturbations, ranging from as large as seed changes to as small as floating-point errors, result in similar levels of variability in generalization (Extended Data Fig.~7), consistent with previous characterizations \cite{summers2021nondeterminism, bhojanapalli2021reproducibility, dodge2020fine, zhuang2022randomness}. These results support riddling as a candidate organizing principle for irreducible model variability in deep learning.

\textbf{Critical learning periods.} Riddling offers a geometric interpretation of the critical learning period \cite{achille2019critical, frankle2020early}, during which sensitivity of the training outcome to perturbations is initially high and then diminishes, but never vanishes \cite{summers2021nondeterminism}.
Geometrically, sensitivity is greater when competing basins occupy a larger fraction of the neighborhood around the network state. In Supplementary Sec.~10, we introduce a sensitivity metric $\bar{s}$ based on this idea and conduct a control study comparing riddled and non-riddled basins. To summarize, a non-riddled basin occupies the entire space near its attractor implying vanishing sensitivity (i.e., $\bar{s} = 0$), contrary to empirical observations \cite{summers2021nondeterminism}. 
Persistent late-training sensitivity (i.e., $\bar{s} \neq 0$) can be explained through riddling, since holes of competing basins exist arbitrarily close to the attractor. 

\textbf{Performance-reproducibility trade-offs.} Riddling-like irreproducibility often coincides with the learning rate that yields peak training performance, suggesting a performance-reproducibility trade-off. 
At smaller learning rates, both parameter- and function-space outcomes become reproducible, but training performance deteriorates.
We identify such trade-offs in the majority of tested configurations (Extended Data Fig.~6 and 8, and Supplementary Sec.~8).
Similar trade-offs exist in the edge-of-stability \cite{cohen2021gradient} and chaotic training \cite{herrmann2022chaotic} literature, where suppressing local instability impairs training performance. The riddling framework extends these perspectives by showing that the relevant instability is not limited to local, trajectory-level sensitivity, but can also occur at the level of global functional outcomes.  

\section*{Discussion}\label{sec4}

In this work, we have investigated the reproducibility of deep learning and proposed riddled basin geometry as a unifying dynamical-systems mechanism underlying its limits. 
In a minimal model, we have shown that riddling emerges from widespread features of deep learning---symmetry \cite{ziyin2025parameter}, attraction to invariant subspaces \cite{chen2023stochastic}, chaos and instability \cite{kong2020stochasticity, herrmann2022chaotic, ly2025optimization}---and produce outcome-level unpredictability, quantified by a near-zero uncertainty exponent, $\phi_p$. 
With $\phi_p \approx 0$, increasing initialization precision does not meaningfully improve reproducibility.
The same signatures arise in the parameter space of deep networks, and an analogous functional exponent, $\phi_f$, diagnoses reproducibility limits in function space, extending the framework to settings where training destinations cannot be explicitly identified. 
The riddling framework thus rationalizes long-standing observations of irreproducibility \cite{gundersen2022sources, bhojanapalli2021reproducibility, ahn2022reproducibility, zhuang2022randomness, dodge2020fine, jiang2021churn, yuan2025give, summers2021nondeterminism, sun2022surprising} and elevates unpredictability to be central for understanding deep learning, in parallel with other physical and computational systems \cite{sommerer1993physical, moore1990unpredictability, bennett1990undecidable, cubitt2015undecidability}. 

Riddled basin geometry arises in a wide range of dynamical systems \cite{sommerer1993physical, aguirre2009fractal}, and local riddling, a weaker property, was previously reported in simple networks trained on chaotic time series \cite{nakajima1996riddled}. We expect riddling could be widespread because its ingredients are prevalent in deep learning. 
Even when the conditions for riddling are only partially met, riddled-like behaviors can persist \cite{pecora1990synchronization, ashwin1994bubbling, lai1999riddling, lai2001pseudo, woltering2000riddled}. Indeed, in the minimal model, we have observed a sequence of phase transitions as the learning rate increases, from no riddling to pseudo-riddling (a mixture of riddled and non-riddled components), true riddling, and transient riddling (Extended Data Fig.~9 and Supplementary Sec.~1). Empirically, we have found riddling-like reproducibility limits across a range of test configurations, including VGGNet, ResNet and BERT trained with different optimizers, regularizers and activation functions (Supplementary Sec.~8). 
Together with related observations of training instability in diverse settings, including large language models and recurrent neural networks \cite{dodge2020fine, yuan2025give, summers2021nondeterminism}, these findings support riddling as a potential unifying mechanism for irreproducibility in deep learning.

The riddling framework complements Edge-of-Stability (EoS) viewpoints \cite{cohen2021gradient, ly2025optimization, sun2022surprising} by addressing the global organization of training outcomes rather than only local trajectory-level instability. 
EoS involves instability along high-curvature directions and provides a route to weakened transverse stability or chaotic dynamics (see Supplementary Sec.~11). 
However, riddling additionally requires competing attractors in symmetry-induced invariant subspaces. These additional ingredients produce qualitative unpredictability \cite{sommerer1993physical}, which is stronger than what can be understood through EoS phenomena alone. 
The riddling perspective thus yields a distinct prediction: systems exhibiting similar EoS behavior may nevertheless differ in reproducibility due to differences in their invariant-subspace or attractor structure. This prediction motivates interventions beyond simply reducing the learning rate, including symmetry-breaking strategies. 
Consistent with this prediction, symmetry removal can mitigate the initialization-dependence of training outcomes \cite{ziyin2025remove}, although the resulting basin geometry remains to be characterized, as destinations are no longer identifiable through matched or vanishing neurons.

Our results further suggest a trade-off between training performance and reproducibility: near-zero parameter- and function-space uncertainty exponents often coincide with the optimal learning rate (Supplementary Sec.~8).
What factors determine whether performance and irreproducibility coincide? Can riddling be analytically connected to generalization bounds, for example through algorithmic stability \cite{chen2018stability}? In Supplementary Sec.~6, we identify trends in this direction: in the riddling-like regime, the generalization error is strongly correlated with the functional difference between networks trained from $\varepsilon$-close initializations, independent of training details. Outside this regime, by contrast, functional differences are near zero and uncorrelated with generalization error. Given the conventional view that generalization is related to the flatness of a solution \cite{hochreiter1997flat}, the relationship between irreproducibility and the structure of the loss landscape warrants further investigation.

As deep learning advances into domains where reproducibility is vital, significant effort has been dedicated to managing the variability caused by non-deterministic software and hardware \cite{milani2016launch, beam2020challenges, liu2021reproducibility, pineau2021improving, bhojanapalli2021reproducibility, summers2021nondeterminism, yuan2025give, jiang2021churn, ahn2022reproducibility, zhuang2022randomness, gundersen2022sources}. Yet, the mechanism by which small perturbations yield large network differences remained unexplored. 
Early work showed that back-propagation is sensitive to initial conditions \cite{kolen1990back}, with convergence-rate maps exhibiting fractal structures, as also found recently in hyperparameter space \cite{sohl2024boundary} (see Supplementary Sec.~3 for connections between parameter-space riddling and hyperparameter-space fractality resembling lakes of Wada \cite{kennedy1991basins}). 
Those studies, however, concerned ordinary ``skinny'' fractals with zero Lebesgue measure. 
In contrast, our analysis of destination maps uncovers ``fat'' fractals, in the form of riddled basins, with positive Lebesgue measure \cite{umberger1985fat}: the basin of attraction for one destination is densely intertwined with points belonging to others. 
Unlike skinny fractality, which degrades reproducibility only near basin boundaries and can be overcome by increasing precision, fat fractality sets a scale-insensitive limit: an arbitrarily small perturbation to any initialization can change the basin in which it resides, and no increase in precision offers recourse. 
Our findings are thus consistent with the interpretation that irreproducibility in deep learning arises not from perturbations per se, but from their coupling with riddled basin geometry.

More broadly, riddling entails a form of unpredictable dynamics that is qualitatively new to neural network training, and in complexity, exceeds the chaotic behaviors studied previously \cite{herrmann2022chaotic, ly2025optimization, kong2020stochasticity}. In idealized dynamical systems with riddled or intermingled basins, related undecidability phenomena can arise, and similar behavior was recently demonstrated in quantum many-body systems with phase diagrams exhibiting fractal structure \cite{cubitt2015undecidability, bausch2021uncomputability, watson2022uncomputably}. Our results show a practical obstruction to robust prediction under finite perturbations, consistent with this dynamical-systems picture.
Deep learning thus joins this broader context, with riddled basin geometry emerging as a candidate organizing principle for irreproducibility in modern artificial intelligence.

\section*{Methods}\label{sec11}

\subsection*{A dynamical systems primer}

Here we define concepts from dynamical systems theory that are relevant for riddled basins. Riddling requires a measure-theoretical definition of an attractor \cite{milnor1985concept}. Consider a dynamical system whose time dependence is generated by the iteration of a map $\Phi: \mathbb{R}^d \rightarrow \mathbb{R}^d$. Suppose there exists an invariant set $A \subset \mathbb{R}^d$, such that $\Phi^t(A) \subset A$ for all $t$. The basin of attraction of $A$ is $\beta(A) \coloneqq \{\bm\uptheta \in \mathbb{R}^d \mid \omega(\bm\uptheta) \subset A\}$, where the $\omega$-limit set $\omega(\bm\uptheta) \coloneqq \cap_{t > 0} \overline{\{\Phi^s (\bm\uptheta_0) \mid s > t\}}$ is the set of accumulation points of the trajectory $\Phi^t(\bm\uptheta_0)$ as $t \rightarrow \infty$. That is, $\beta(A)$ contains all points that accumulate in $A$. The invariant set $A$ is a Milnor attractor if $l(\beta(A)) > 0$ and $l(\beta(A) \setminus \beta(A')) > 0$ for any proper compact subset $A' \subset A$, where $l$ is the $d$-dimensional Lebesgue measure. In contrast to other definitions of an attractor, the basin of a Milnor attractor need not contain an open neighborhood of the attractor itself. Milnor attractors are essential for the formulation of riddled basins, as riddling can occur even arbitrarily close to the attractor, violating classical notions of Lyapunov stability. A basin of attraction $\beta(A)$ is riddled if for all $\bm\uptheta \in \beta(A)$ and $\delta > 0$ we have $l(B_{\delta}(\bm\uptheta) \cap \beta(A)^{\mathsf{c}}) > 0$, where $B_\delta(\bm\uptheta)$ denotes the $\delta$-neighborhood of $\bm\uptheta$ \cite{alexander1992riddled}. Thus any initialization attracted to $A$ is arbitrarily close to another initialization leading to a different attractor.

As established in the dynamical systems theory, the theoretical conditions sufficient for riddling are: the existence of a chaotic attractor within an invariant subspace, containing a zero measure set of transversely unstable periodic points, and at least one competing attractor elsewhere. We now clarify these conditions: A subspace $\mathcal{P}$ is invariant if trajectories originating within the subspace remain in it indefinitely (i.e., $\Phi(\mathcal{P}) \subset \mathcal{P}$). Within the invariant subspace is a chaotic attractor $A$, meaning that there exists a basin $\beta(A)$ of initializations off the subspace that are drawn towards $A$. Under the classical definition of chaos, a chaotic attractor contains periodic points; a point is periodic if by taking a finite number of steps from it, the system would return to the point (i.e., $\Phi^m(\bm{\uptheta}) = \bm{\uptheta}$). We regard periodic points in the chaotic attractor as atypical as they constitute a set of measure zero. These periodic points are transversely unstable if small transverse perturbations from them grow over time. Through local stability analysis \cite{herrmann2022chaotic}, the transverse instability of a point can occur if in its vicinity the curvature of the loss function in the transverse direction is either negative or large relative to the learning rate; the latter is related to the edge of stability phenomenon \cite{cohen2021gradient} (see Supplementary Sec.~11 for experimental evidence). 

We now illustrate how these conditions constitute the standard dynamical mechanism for riddling (see e.g., \cite{ott1994transition} for a classical example). The chaotic attractor within the invariant subspace, $A \subset \mathcal{P}$, has a basin of attraction $\beta(A)$, so a positive volume of initial conditions off the subspace converges towards $A$. However, the transverse instability of a periodic point $P$ embedded in $A$ implies that, simultaneously, there are initializations arbitrarily close to the invariant subspace that are repelled away from it. The presence of competing attractors then converts these ejected trajectories into different long-term outcomes. In other words, such ejected trajectories may follow heteroclinic connections towards a competing attractor, for example one located in another invariant subspace, like $\mathcal{P}_-$ in the minimal model, or at infinity. If $P$ also has stable directions, this trajectory aligns with a stable manifold funneling nearby points towards the same destination. The set of initial conditions that converge to the competing attractor can thus be visualized as a ``hyperwedge'' anchored at the periodic point (see the schematic in Extended Data Fig.~2). Because pre-images of these points also evolve towards the same competing attractor, the hyperwedge structure is reproduced at the pre-iterates of $P$. Owing to the chaotic dynamics on $A$, these pre-iterates are dense in the attractor. Consequently, a dense collection of hyperwedges is generated throughout the basin. These hyperwedges perforate $\beta(A)$, producing the characteristic fat-fractal geometry of a basin that is densely riddled with initial conditions leading elsewhere.

\subsection*{Chaotic attractor visualization}
To visualize the chaotic attractor in the permutation-invariant subspace $\mathcal{P}_{+}$, we train the network for $10^5$ epochs from a random initialization inside $\mathcal{P}_{+}$. We visualize each iterate of the trajectory. The qualitative nature of this image is independent of the initialization.

\subsection*{Lyapunov exponent calculation}
The Lyapunov spectrum provides a quantitative diagnostic to determine the nature of stability of an attractor. We briefly recall notions surrounding its definition and measurement. The Jacobian matrix of the discrete-time map $\Phi$ is $\mathbf{J}(\bm\uptheta) = \partial \Phi (\bm\uptheta)/\partial \bm\uptheta$ for any $\bm\uptheta \in \mathbb{R}^d$. The spectrum of (infinite-time) Lyapunov exponents is given by $\lambda_i = \ln \mu_i$, where $\mu_i$ are the eigenvalues of the Lyapunov matrix $\mathbf{\Lambda} \coloneqq \lim_{t \rightarrow \infty} [\mathbf{Y}_t^\top \mathbf{Y}_t]^{1/2t}$, and $\mathbf{Y}_t = \mathbf{J}(\bm\uptheta_{t-1})\mathbf{J}(\bm\uptheta_{t-2})\dots \mathbf{J}(\bm\uptheta_{0})$ is the Jacobian matrix of the $t$-times iterated map. The finite-time Lyapunov exponents $\lambda_i^T$ are defined analogously without the limit, using the eigenvalues of $[\mathbf{Y}^\top_T \mathbf{Y}_T]^{1/2T}$. While the finite-time values fluctuate with the initialization $\bm\uptheta_0$, the infinite-time limit is almost surely independent of it with respect to the natural ergodic measure on an attractor $A$, according to Oseledets theorem. Since the direct computation of the Lyapunov matrix is numerically unstable, an othornormalization scheme is used in practice. 

We apply the treppen-iteration algorithm \cite{eckmann1985ergodic}:
\begin{equation}
    \mathbf{J}(\bm\uptheta_{j-1})\mathbf{Q}^{j-1} = \mathbf{Q}^{j}\mathbf{R}^{j-1},
\end{equation}
where the right-hand side is obtained through the QR decomposition of the left-hand side. Here $\mathbf{J}(\bm\uptheta) = \mathbf{I} - \eta \mathbf{H}(\bm\uptheta)$ is the Jacobian matrix for deterministic gradient descent, $\mathbf{I}$ is the identity matrix, $\mathbf{H}$ is the Hessian matrix of the loss function $L$, $\mathbf{R}^j$ is an upper triangular matrix and $\mathbf{Q}^j$ is an orthonormal matrix whose initial value at $j=0$ can be chosen arbitrarily. The Lyapunov exponents are then given by:
\begin{equation}
    \lambda_i = \lim_{t \rightarrow \infty} \frac{1}{t} \sum_{j=0}^{t-1} \ln |R_{ii}^j|,
\end{equation}
where $R_{ii}^j$ denotes the $i$-th diagonal element of $\mathbf{R}^j$. Accordingly, the finite-time Lyapunov exponents are given by:
\begin{equation}
    \lambda_i^T = \frac{1}{T} \sum_{j=0}^{T-1} \ln |R_{ii}^j|.
\end{equation}

For an initialization within an invariant subspace, $\bm\uptheta_0 \in \mathcal{P}$, the Lyapunov exponents can be partitioned into two sets corresponding to either longitudinal or transverse expansion \cite{alexander1992riddled}. They contain $d_{\mathcal{P}}$ and $d - d_{\mathcal{P}}$ exponents, respectively, where $d_{\mathcal{P}}$ denotes the dimension of the invariant subspace $\mathcal{P}$. To calculate these exponents, we define the basis matrix
\begin{equation}
    \mathbf{E} = \frac{1}{\sqrt{2}} \begin{pmatrix} 1 & 0 & 1 & 0 \\ 1 & 0 & -1 & 0 \\ 0 & 1 & 0 & 1 \\ 0 & 1 & 0 & -1 \end{pmatrix} = \begin{pmatrix} \mathbf{e}_1 & \mathbf{e}_2 & \mathbf{e}_3 & \mathbf{e}_4 \end{pmatrix}. \label{eq:basis}
\end{equation}
Additionally, we construct $\mathbf{E}_{\parallel} = \begin{pmatrix} \mathbf{e}_1 & \mathbf{e}_2 \end{pmatrix}$ and $\mathbf{E}_{\perp} = \begin{pmatrix} \mathbf{e}_3 & \mathbf{e}_4 \end{pmatrix}$, which span $\mathcal{P}$ and its orthogonal complement, respectively. Since the subspace $\mathcal{P}$ is invariant, for any $\bm\uptheta \in \mathcal{P}$ and $\mathbf{u} \parallel \mathcal{P}$ we have $\Phi(\bm\uptheta + \mathbf{u}) \in \mathcal{P}$. By linearization, $\mathbf{J}(\bm\uptheta) \mathbf{u} \approx \Phi(\bm\uptheta + \mathbf{u}) - \Phi(\bm\uptheta) \parallel \mathcal{P}$. Thus, $\mathbf{E}_{\perp}^\top \mathbf{J}(\bm\uptheta) \mathbf{E}_{\parallel} = \mathbf{0}$. For $L \in C^2$, the Jacobian is symmetric with block-diagonal representation,
\begin{equation}
    \mathbf{E}^\top \mathbf{J}(\bm\uptheta) \mathbf{E} = \begin{pmatrix}
        \mathbf{J}_{\parallel}(\bm\uptheta) & \mathbf{0}\\
        \mathbf{0} & \mathbf{J}_{\perp}(\bm\uptheta)
    \end{pmatrix},
\end{equation}
where $\mathbf{E} = \begin{pmatrix} \mathbf{E}_{\parallel} & \mathbf{E}_{\perp}\end{pmatrix}$, $\mathbf{J}_{\parallel}(\bm\uptheta) = \mathbf{E}_{\parallel}^\top \mathbf{J}(\bm\uptheta) \mathbf{E}_{\parallel}$ and $\mathbf{J}_{\perp}(\bm\uptheta) = \mathbf{E}_{\perp}^\top \mathbf{J}(\bm\uptheta) \mathbf{E}_{\perp}$. It follows that $\mathbf{E}^\top \mathbf{Y}_t \mathbf{E}$ and $\mathbf{E}^\top \mathbf{\Lambda} \mathbf{E}$ have block-diagonal representations, such that the eigenvalues of $\mathbf{\Lambda}$ are the union of the eigenvalues of $\mathbf{\Lambda}_\parallel = \mathbf{E}_{\parallel}^\top \mathbf{\Lambda} \mathbf{E}_{\parallel}$ and $\mathbf{\Lambda}_\perp = \mathbf{E}_{\perp}^\top \mathbf{\Lambda} \mathbf{E}_{\perp}$. Therefore, the longitudinal and transverse Lyapunov exponents can be calculated by independently using the QR algorithm on their respective blocks:
\begin{equation}
    \mathbf{J}_{\parallel}(\bm\uptheta_{j-1})\mathbf{Q}_{\parallel}^{j-1} = \mathbf{Q}_{\parallel}^{j}\mathbf{R}_{\parallel}^{j-1} \quad \text{and} \quad \mathbf{J}_{\perp}(\bm\uptheta_{j-1})\mathbf{Q}_{\perp}^{j-1} = \mathbf{Q}_{\perp}^{j}\mathbf{R}_{\perp}^{j-1}.
\end{equation}
To determine the Lyapunov exponents of the chaotic attractor in $\mathcal{P}_{+}$, we choose $\mathbf{Q}^0_\parallel = \mathbf{Q}^0_\perp = \mathbf{I}$ and apply the treppen-iteration algorithm for $10^5$ iterations starting from a random initialization $\bm\uptheta_0 \in \mathcal{P}_{+}$. In Supplementary Sec.~4, we present numerical analysis of local Lyapunov exponents in deep neural network training.

\subsection*{Deep neural network training configuration}
Because a rigorous characterization of riddling requires the simultaneous training of tens of thousands of networks, it is computationally infeasible to use state-of-the-art architectures and learning tasks at full scale. Accordingly, we design a training configuration that is realistic in the sense that we maximize the complexity of the architecture and learning task under constrained compute (e.g., experiments are replicable within one week on a high performance computing cluster). We use a VGG-12 network, which is a 12-layer implementation of the VGG architecture that is widely used for image classification \cite{simonyan2014very}. The VGG-12 comprises 9 convolutional layers and 3 fully-connected layers. To facilitate faster training, we taper the network to \num[group-separator={,}]{12036} parameters by reducing the number of channels per layer and removing biases. Bias removal also induces a symmetry that we exploit to render higher resolution visualizations (see the following sections). Networks are trained to perform image classification on the MNIST dataset corrupted with $50\%$ label noise (i.e., half of the training data is intentionally mislabeled). We note that there is a vast literature devoted to label noise as it is a ubiquitous issue in practical machine learning \cite{frenay2013classification}. Here we introduce label noise because it increases task difficulty while accelerating stochastic collapse \cite{chen2023stochastic}, reducing the duration of our experiments; note that riddling still occurs without label noise (see Extended Data Fig.~5 and Supplementary Sec.~8). 

Training minimizes the cross-entropy loss using stochastic gradient descent with learning rate $\eta = 0.1$, batch size $b=128$, momentum 0.9, and weight decay $5 \times 10^{-4}$. Despite $50\%$ label noise, the random Kaiming-initialized \cite{he2015delving} network in Fig.~\ref{fig:4}\hyperref[fig:4]{a} achieves a testing accuracy of $97.93\%$ (see Extended Data Fig.~10). Such high accuracy, although initially surprising perhaps, is not unusual considering previous empirical results: a simple convolution network can achieve approximately 90\% test accuracy with over 99\% label noise \cite{rolnick2017deep}. Evidently, our experimental design fulfills a central desideratum of practical deep learning, that is generalization. 

In all experiments, we fix mini-batch ordering to eliminate non-determinism associated with the training algorithm. When coupled with CPU computation, this ensures determinism: all initializations are described by the same dynamical system, which yields the same outcome upon repeated simulation. For uncertainty exponent calculations, which test the stability of training outcomes under small perturbations, determinism controls for the accumulation of random deviations during training (e.g., due to different mini-batches) that could otherwise amount to outcome-level differences between runs. This isolates complex deterministic dynamics, such as riddling, as the mechanism explaining parameter-space irreproducibility. Note that SGD with fixed mini-batch order, while deterministic, is still periodically forced due to rotating mini-batches that can lead to complex stochastic-like behavior; we find that riddling-like behavior persists even without such forcing in full-batch gradient descent (see Supplementary Sec.~8).

\subsection*{Imaging basins of attraction}

In Fig.~\ref{fig:3}, we train each minimal neural network model for $10^3$ epochs. We consider an initialization to be convergent to $\mathcal{P}_{\pm}$ if $\bm\uptheta_t$ remains finite and $d_{\pm}(\bm\uptheta_{1000}) < D$, where $d_{\pm}(\bm\uptheta) = \Vert \mathbf{w}_1 \mp \mathbf{w}_2 \Vert$ is a distance metric to $\mathcal{P}_{\pm}$ subspaces and $D$ is the proximity threshold. Note that $d_{\pm}(\bm\uptheta) = \sqrt{2} d(\bm\uptheta, \mathcal{P}_{\pm})$, where $d(\theta, \mathcal{P}_{\pm})$ is the Euclidean distance between $\bm\uptheta$ and $\mathcal{P}_{\pm}$. If $\bm\uptheta_t$ diverges or $d_{\pm}(\bm\uptheta_{1000}) \geq D$, we consider the initialization to be convergent to an attractor off the invariant planes. We find that the value of $D$ does not change the qualitative nature of the patterns in the destination map, except when $D$ is too small (e.g., $D \lesssim 10^{-4}$), which produces a speckle pattern of white points. We use $D = 3$; at this value, all white points are en route to infinity, so white approximates the basin of divergence.

In Figs.~\ref{fig:4}\hyperref[fig:4]{b-d} and Fig.~\ref{fig:5}, we train each VGG-12 network for 30 epochs. We identify the parity-invariant subspace $\mathcal{P}^J_0$ according to the index set of its vanishing neurons, $J$. We consider an initialization to be convergent to $\mathcal{P}_0^J$ if the vectorized weights of the neurons in $J$ have Euclidean norms less than $10^{-2}$ at the end of training: $||\mathbf{w}_j|| < 10^{-2}$ for all $j \in J$. Although changes to the destination map are expected with further training (see Supplementary Sec.~9), the accuracy on the testing dataset stabilizes after approximately 30 epochs (see Extended Data Fig.~10 as an example). All initializations in the grid that do not limit to the origin achieve strong generalization with an average test accuracy of $(97.6 \pm 0.5)\%$. Those approaching the origin perform at chance level.

In Fig.~\ref{fig:3}\hyperref[fig:3]{a} and Fig.~\ref{fig:4}\hyperref[fig:4]{b}, we compute the destinations only for initializations in the first quadrant. To obtain the full image, we reflect across the transverse and longitudinal axes. The following section explains the symmetries that enable this shortcut.

\subsection*{Symmetry-induced invariant subspaces}

An affine subspace is invariant when the neural network is reflection-symmetric around it \cite{chen2023stochastic}. Although this symmetry need only be approximate (i.e., in a neighborhood of the subspace), it is exact for the permutation- and parity-invariant subspaces considered in this work. Here we establish three results. 

First, a neural network with an odd activation function $\sigma$ is reflection-symmetric across an affine subspace $\mathcal{P}^J_0$ with set $J$ indexing multiple neurons in the same hidden layer $l$. This generalizes the single-neuron case (i.e., $|J| = 1$) \cite{chen2023stochastic}. A reflection across $\mathcal{P}^J_0$ flips the sign of parameters of neurons in $J$, including the incoming weights $\mathbf{w}^{(l)}$, outgoing weights $\mathbf{w}^{(l+1)}$ and biases $b^{(l)}$ of each neuron. The input to the hidden layer $\mathbf{x}^{(l-1)}$ is unchanged. As a result, their activations $\sigma(\mathbf{w}^{(l)} \cdot \mathbf{x}^{(l-1)} + b^{(l)})$ flip sign (because $\sigma$ is odd). This is canceled by the sign flip of the outgoing weights, leaving the neural network output invariant. Although there exists other parity-invariant subspaces (e.g., sets $J$ indexing neurons of different layers that do not share weights), the result here encompasses the vast majority ($\approx 99.2\%$) of the destinations observed in VGG-12 network training. 

Second, if in addition the neural network has an even number of hidden layers and no bias parameters, then it is reflection-symmetric across the affine subspace $\mathcal{N}^{J}_0$ that is transverse to $\mathcal{P}^J_0$. This subspace corresponds to all parameters not belonging to neurons in set $J$ vanishing. A reflection across $\mathcal{N}^{J}_0$ reverses the sign of these parameters. As a result, a $(-1)$ factor is accrued at each hidden layer before and after the $l$-th hidden layer, as well as the output layer. Note that the $l$-th hidden layer does not contribute a factor. To see this, note that the input to the $l$-th hidden layer changes as $\mathbf{x}^{(l-1)} \mapsto (-1)^{l-1} \mathbf{x}^{(l-1)}$. For neurons in $J$, their activations become $(-1)^{(l-1)} \sigma(\mathbf{w}^{(l)} \cdot \mathbf{x}^{(l-1)})$. For neurons not in $J$, their activations become $(-1)^{(l)} \sigma(\mathbf{w}^{(l)} \cdot \mathbf{x}^{(l-1)})$ since $\mathbf{w}^{(l)} \mapsto -\mathbf{w}^{(l)}$. Another sign flip at the outgoing weights of neurons not in $J$ cancels this additional factor. The total factor accrued at the neural network output is $(-1)^{l_{\text{total}}}$, where $l_{\text{total}}$ is the total number of hidden layers. Thus, a neural network is reflection-symmetric across $\mathcal{N}^{J}_0$ if it has an even number of hidden layers and no bias parameters. Under these conditions, $\mathcal{N}^{J}_0$ is an invariant subspace \cite{chen2023stochastic}. Since the VGG-12 network satisfies these conditions, the result implies the left-right symmetry of Fig.~\ref{fig:4}\hyperref[fig:4]{b-c}.

Third, a neural network with an odd activation is reflection-symmetric across additional subspaces $\mathcal{P}^{i,j}_- \coloneqq \{\bm\uptheta \in \mathbb{R}^d \mid \mathbf{w}_i = - \mathbf{w}_j\}$, where $\mathbf{w}_i$ and $\mathbf{w}_j$ are the vectorized parameters of the $i$ and $j$ neurons in the same hidden layer. Reflection across this subspace corresponds to the composition of a permutation of parameters and a sign flip. The neural network is invariant under permutations, which generates the permutation-invariant subspaces $\mathcal{P}^{i,j}_+ \coloneqq \{\bm\uptheta \in \mathbb{R}^d \mid \mathbf{w}_i = \mathbf{w}_j\}$ \cite{chen2023stochastic}. As shown above, a neural network is also invariant under the sign reversal of neurons in the same hidden layer. Taken together, the neural network is invariant under the composition. The result indicates that $\mathcal{P}_- \equiv \mathcal{P}^{1,2}_{-}$ is an invariant subspace of the minimal model, and explains the left-right symmetry in Fig.~\ref{fig:3}\hyperref[fig:3]{a}.

\subsection*{Uncertainty exponent calculation}

The fraction of $\varepsilon$-uncertain initializations, $f(\varepsilon)$, in a small region depends on that region's distance from the relevant attractor. Thus, we must calculate $f(\varepsilon)$ using initializations that are a fixed distance away. 

For the minimal model (Fig.~\ref{fig:3}), we consider initializations equidistant from the two permutation-invariant subspaces $\mathcal{P}_{\pm}$, such that $d_{\pm}^2(\bm\uptheta) = 8$. The initializations that satisfy this constraint can be expressed as $\bm\uptheta = a_1 \mathbf{e}_{1} + a_2 \mathbf{e}_{2} + a_3 \mathbf{e}_{3} + a_4 \mathbf{e}_{4}$, where the orthonormal basis vector $\mathbf{e}_i$ is the $i$-th column vector of the matrix in equation~(\ref{eq:basis}) and $a_1^2 + a_2^2 = a_3^2 + a_4^2 = 4$. Thus, we randomly generate initializations by sampling coefficients from two separate circles. For every such initialization, we generate another by perturbing each network parameter with a uniform random value $U(-\varepsilon, \varepsilon)$, where $\varepsilon$ is the uncertainty. For each uncertainty value, we train $n=10^4$ pairs of initializations for $10^3$ epochs and calculate $f(\varepsilon)$ as the fraction of pairs whose training outcome differ. We compute the standard error of $f(\varepsilon)$ as $\sqrt{f(\varepsilon)(1-f(\varepsilon))/n}$.

Because the training of the VGG-12 network has many symmetry-induced invariant subspaces, specifying the distance of initializations from each invariant subspace is intractable. Instead, we generate initializations by perturbing the parameters of a fixed reference initialization by uniform random values, distributed as $U(-\varepsilon/2, \varepsilon/2)$, so that the maximum possible separation between any pair of initializations is $\varepsilon$. We compute $f(\varepsilon)$ and its standard error by bootstrap resampling. Unlike the minimal model, this is possible here because initializations are independently sampled within the same region for each uncertainty value $\varepsilon$. Specifically, for each $\varepsilon$ we determine the destination of $10^3$ initializations. We randomly pair initializations to obtain a bootstrap sample of $f(\varepsilon)$. We estimate the standard error from the bootstrap distribution. Computational constraints aside, this procedure can be straightforwardly applied to arbitrary architectures, including state-of-the-art networks, if the set of symmetry-induced invariant subspaces are known; see Supplementary Sec.~8 for additional examples. The uncertainty exponent calculation in the schematic (Fig.~\ref{fig:1}) also applies this procedure, except with $10^4$ pairs of initializations for each $\varepsilon$. In this case, the fixed reference initialization is $(\bm\uptheta \cdot \mathbf{e}_{\parallel}, \bm\uptheta \cdot \mathbf{e}_{\perp}) = (0.539, 1.819)$, which is marked by the white cross in Fig.~\ref{fig:1}. 

For deep neural networks, we also calculate the functional uncertainty exponent, which is of practical relevance. This exponent is obtained by fitting a power law to a functional similarity measure. Although we present results for churn here, the same analysis can be performed with other measures of functional similarity (see Extended Data Fig.~3 and Supplementary Sec.~5). In Fig.~\ref{fig:5}\hyperref[fig:5]{e}, we calculate the mean and standard deviation of churn across $n=250$ $\varepsilon$-close pairs of initializations generated randomly through the same perturbation method as above. That is, each of the two initializations in the pair are obtained by perturbing the parameters of a fixed reference initialization with uniform random values distributed as $U(-\varepsilon/2, \varepsilon/2)$. 

\vspace{.5em}

\noindent \textbf{Data availability} All referenced datasets are publicly available: MNIST \cite{lecun1998gradient} (\url{http://yann.lecun.com/exdb/mnist/}), CIFAR-10 \cite{krizhevsky2009learning} (\url{https://www.cs.toronto.edu/~kriz/cifar.html}), IMDB (\url{https://huggingface.co/datasets/stanfordnlp/imdb}). All primary data generated during the study are available on Code Ocean under \url{https://doi.org/10.24433/CO.8310374.v1} \cite{bacc5f84-2472-4712-9755-3972ca74a072}. Source data are available with this paper.

\vspace{.5em}

\noindent \textbf{Code availability} The code for simulations and analyses of riddled basin geometry in neural network training is available without restrictions on Github (\url{https://github.com/anly2178/riddled_basins_neural_network}) and Code Ocean under \url{https://doi.org/10.24433/CO.8310374.v1} \cite{bacc5f84-2472-4712-9755-3972ca74a072}.

\bibliographystyle{unsrtnat-initials}
\bibliography{bibliography}

\vspace{.5em}

\noindent \textbf{Acknowledgements}
This work was supported by the Australian Research Council (grant no. DP160104368).

\vspace{.5em}

\noindent \textbf{Author Contributions} A.L. and P.G. designed the study, performed the research and wrote the paper.

\newpage
\onecolumngrid

\renewcommand{\figurename}{Extended Data Fig.}
\counterwithin*{figure}{part}
\stepcounter{part}

\begin{figure}[h]
    \centering
    \includegraphics{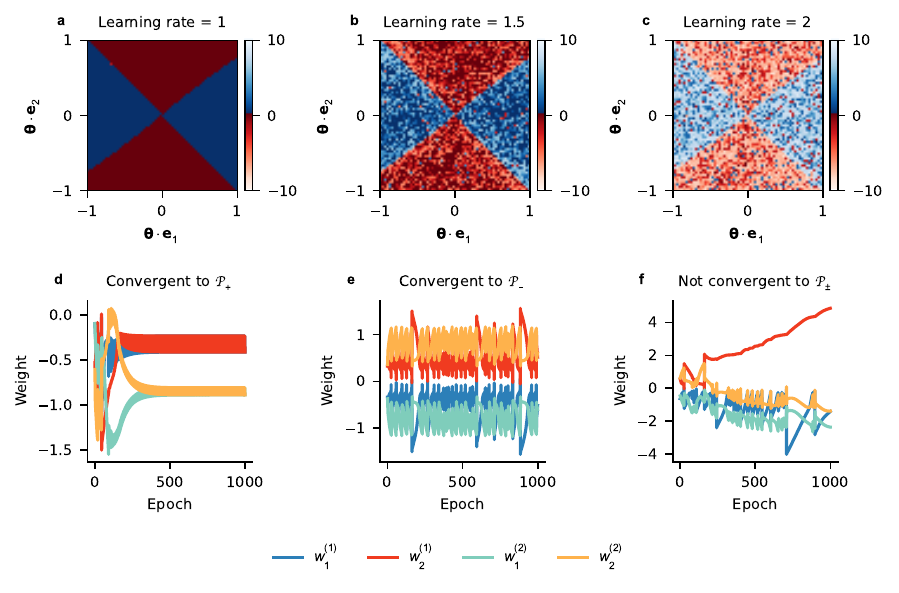}
    \caption{\textbf{Training of the minimal model converges to permutation-invariant planes.} A $64 \times 64$ uniform grid of initializations on the plane spanned by random orthonormal vectors, $\mathbf{e}_1$ and $\mathbf{e}_2$, is trained for $10^3$ epochs. \textbf{a,} Training with a learning rate of $\eta = 1$. Color encodes the nearest invariant subspace: $\mathcal{P}_{+}$ is blue and $\mathcal{P}_{-}$ is red. Color intensity represents the distance to this subspace, $d_{\pm}(\bm\uptheta) = \Vert \mathbf{w}_1 \mp \mathbf{w}_2 \Vert^2$. \textbf{b,} Same as (\textbf{a}), with $\eta = 1.5$. \textbf{c,} Same as (\textbf{a}), with $\eta = 2$. \textbf{d,} Evolution of the weights for a representative initialization from the $\eta=1.5$ grid that converges to $\mathcal{P}_{+}$. \textbf{e,} Same as (\textbf{c}), for an initialization that converges to $\mathcal{P}_{-}$. \textbf{f,} Same as (\textbf{c}), for an initialization that does not converge to either $\mathcal{P}_{+}$ or $\mathcal{P}_{-}$.}
    \label{fig:invariant}
\end{figure}

\begin{figure}[h]
    \centering
    \includegraphics{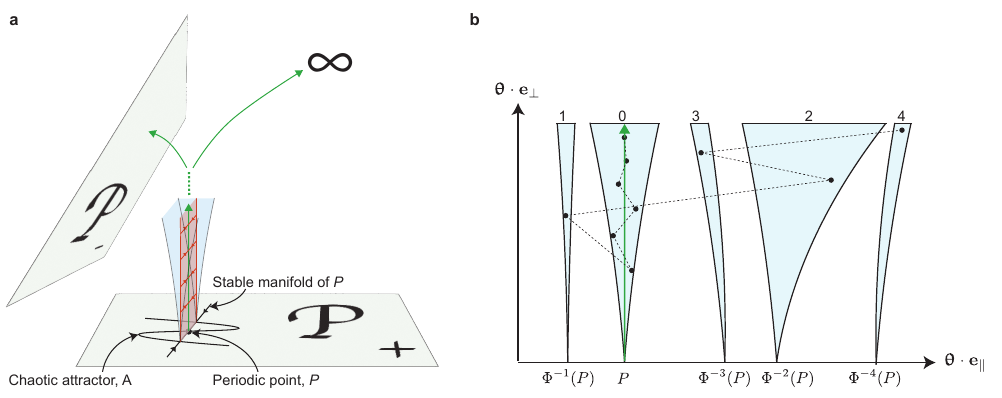}
    \caption{\textbf{Schematic of the mechanism for riddling.} \textbf{a,} A transversely unstable periodic point $P$ is embedded in the chaotic attractor $A \subset \mathcal{P}_+$. Around the stable manifold (orange area) of a heteroclinic trajectory (green arrows), there exists a hyperwedge of initializations (blue volume) whose orbits leave $A$ and either converge to the attractor in $\mathcal{P}_-$ or diverge to infinity. \textbf{b,} The hyperwedge anchored at the periodic point $P$ is labeled 0. Initializations within its pre-images are also ejected away from $\mathcal{P}_+$. For example, the dots and dashed line depicts the trajectory of an initialization starting within the fourth pre-image of the hyperwedge (labeled 4); it travels across the pre-images until it reaches the original hyperwedge, where it follows the heteroclinic trajectory to the competing outcome. These pre-images constitute a dense set of hyperwedges anchored at the pre-iterates of $P$, denoted $\Phi^{-m}(P)$, which are dense in the attractor $A$.}
    \label{fig:mechanism}
\end{figure}

\begin{figure}[h]
    \centering
    \includegraphics{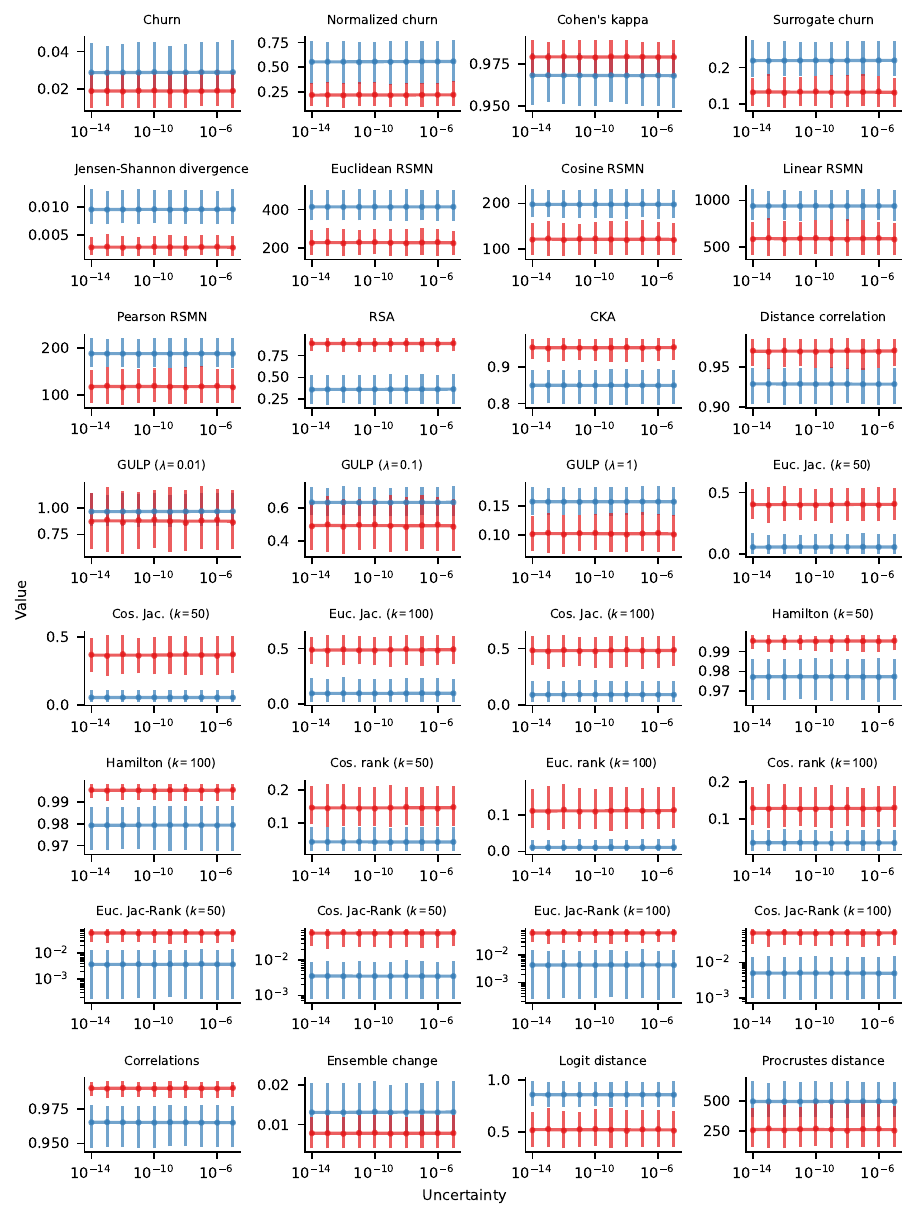}
    \caption{\textbf{Scale-insensitive functional similarity measures.} Each of the 32 measures defined in Supplementary Sec.~5 is evaluated within the riddling-like regime (blue, $\eta = 0.1$) and outside of it (red, $\eta = 0.001$). Dots and error bars denote the mean and 2.5-to-97.5 percentile range computed from $n=250$ $\varepsilon$-close pairs for each uncertainty scale $\varepsilon$. Outside the riddling-like regime, there are substantially smaller functional and representational differences across all measures, indicating markedly higher reproducibility (see Supplementary Sec.~6 for details).}
    \label{fig:sensitivity}
\end{figure}

\begin{figure}[h]
    \centering
    \includegraphics{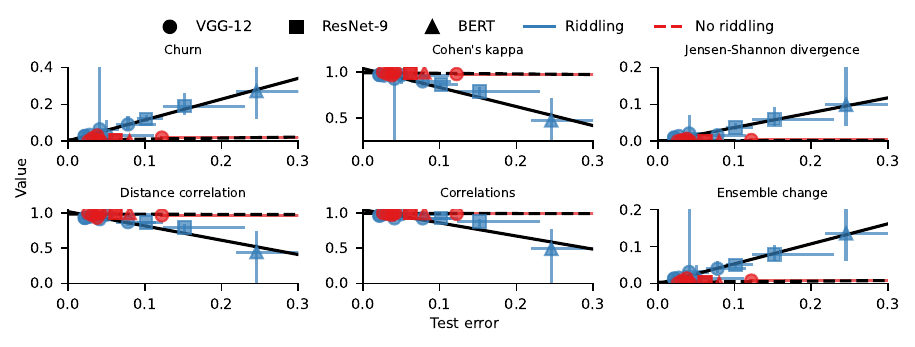}
    \caption{\textbf{Diagnostic for riddling-like behaviors based on functional similarity measures.} Mean functional similarity is visualized as a function of the mean test error within the riddling-like regime (blue, labeled ``riddling'') and outside of it (red, labeled ``no riddling'') across all experimental configurations. Each experimental configuration is labeled by a code: VGG-main, VGG-nln, VGG-bias, VGG-ReLU, Res-SGD, Res-AdamW, BERT (see Supplementary Sec.~8 for details of these experiments). Error bars denote the 2.5-to-97.5 percentile range for both similarity values and test errors. The sample size is $n=50$ for VGG-bias, Res-SGD, BERT, Res-AdamW; $n=75$ for VGG-GD; $n=250$ for VGG-main, VGG-nln, VGG-ReLU, VGG-L1. Error bars exceeding the axis limits are clipped for visual clarity. In the riddling-like regime, functional similarity and generalization exhibit a linear trend (solid black line). Outside this regime, functional differences are negligible, independent of generalization (dashed black line). See Supplementary Sec.~6 for further details, including fitted parameters.}
    \label{fig:similarity_trends}
\end{figure}

\begin{figure}[h]
    \centering
    \includegraphics[width=\textwidth]{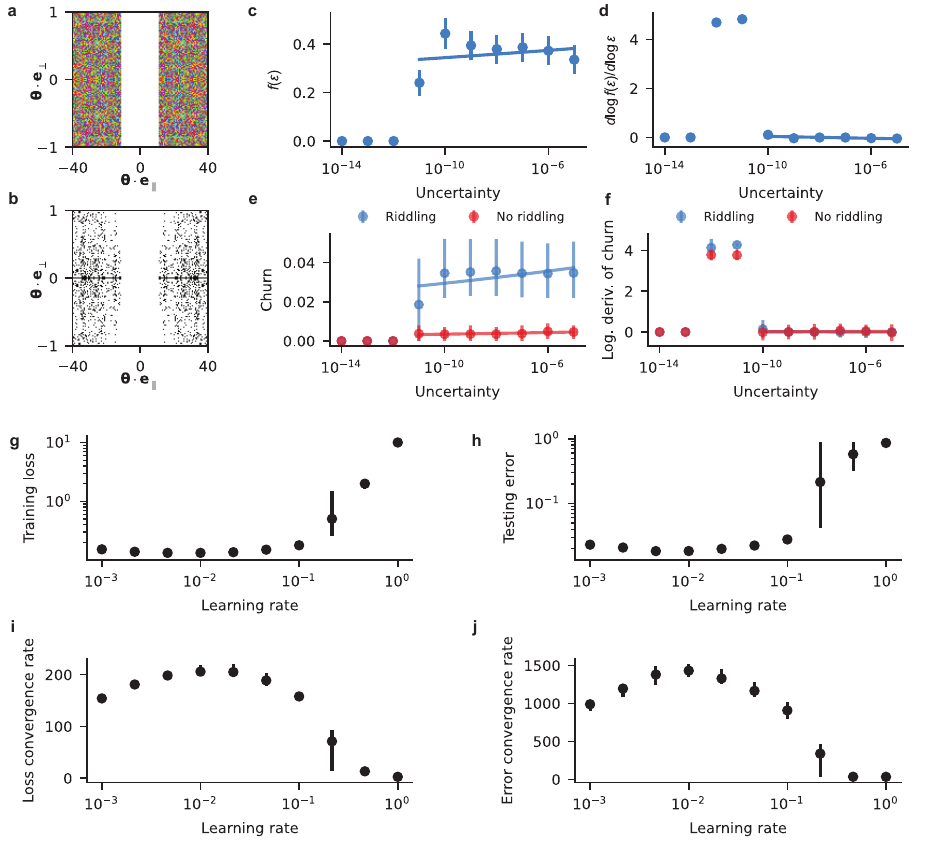}
    \caption{\textbf{Riddling analysis for training without label noise.} Riddling analysis is applied to a VGG-12 network, identical to the one used in Figs.~\ref{fig:4} and \ref{fig:5}, trained to classify MNIST without label noise; see Supplementary Sec.~8 for full details. \textbf{a,} Parameter-space destination map on a $127 \times 127$ grid. \textbf{b,} Same as (\textbf{a}), except initializations converging to the permutation-invariant subspace at $\bm{\uptheta} \cdot \mathbf{e}_\perp = 0$ are colored black. All other destinations are white. \textbf{c,} The uncertainty fraction $f(\varepsilon)$ for initializations within a $\varepsilon$-hypercube centered at $(\bm{\uptheta}\cdot \mathbf{e}_{\parallel}, \bm{\uptheta}\cdot \mathbf{e}_{\perp}) = (26, 0.1)$. Networks are trained on CPU. Dots and error bars represent the mean and 95\% confidence interval across $n=250$ $\varepsilon$-close pairs for each uncertainty $\varepsilon$. A power-law fit $f(\varepsilon) \propto \varepsilon^{\phi_p}$ (solid line) yields the uncertainty exponent $\phi_p = -0.003 \pm 0.007$. \textbf{d,} The logarithmic derivative of (\textbf{c}). \textbf{e,} Dots and error bars represent the mean churn and 2.5-to-97.5 percentile range across $n=250$ $\varepsilon$-close pairs for each $\varepsilon$ in the riddling-like regime (blue, labeled ``riddling''; $\eta = 0.1$) and outside of it (red, labeled ``no riddling''; $\eta = 0.001$). In the riddling-like regime, the churn is significant ($3.5\%$ on average, before suddenly dropping to zero at small initialization uncertainty due to floating point errors; see Supplementary Sec.~8) and shows little improvement as uncertainty $\varepsilon$ is reduced by 6 orders of magnitude ($\phi_f = 0.02 \pm 0.01$). Outside this regime, the churn is practically insignificant (approximately $0.4\%$). \textbf{f,} The logarithmic derivative of (\textbf{e}). \textbf{g,} Dots and error bars show the mean and full range of training loss across $n=5$ independent 30-epoch training runs. \textbf{h,} Testing error. \textbf{i,} Loss convergence rate, defined as $\sum_t \bar{L}_t^{-1}$ where $\bar{L}_t$ denotes the average training loss in the $t$-th epoch. \textbf{j,} Error convergence rate defined analogously using the testing error. This training configuration does not exhibit a performance-reproducibility trade-off as riddling-like irreproducibility does not coincide with optimal performance.}
    \label{fig:no_label_noise}
\end{figure}

\begin{figure}[h]
    \centering
    \includegraphics{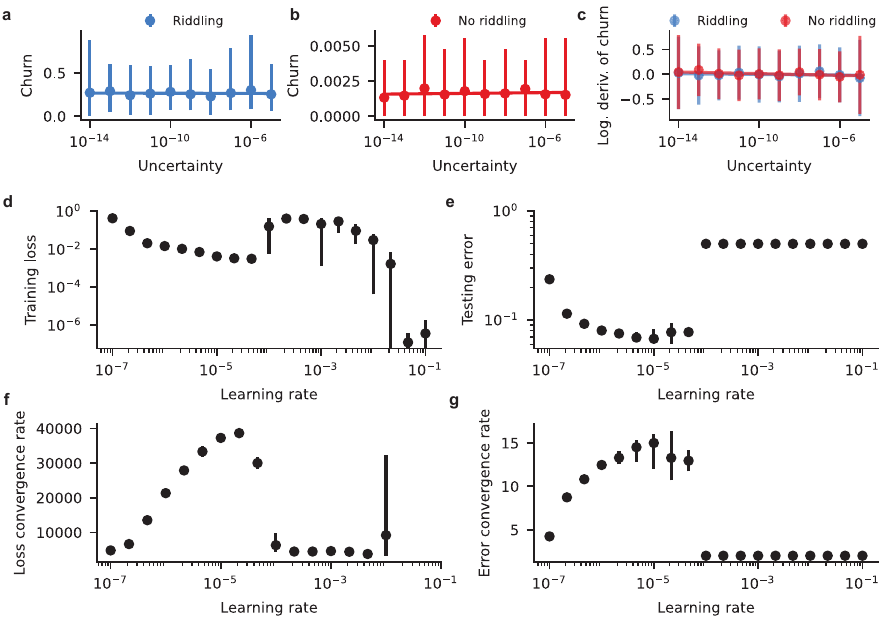}
    \caption{\textbf{Riddling analysis for BERT fine-tuning.} A pre-trained BERT transformer \cite{devlin2019bert} is fine-tuned to perform sentiment classification on the IMDB dataset. Its final state is used as the reference point for functional analysis. See ``Methods'' for further details on the experimental configuration. \textbf{a,} At the near-optimal learning rate of $\eta = 5 \times 10^{-5}$, churn is significant ($27\%$ on average) and shows no improvement as uncertainty $\varepsilon$ is reduced by 10 orders of magnitude ($\phi_f = -0.001 \pm 0.004$), which is evidence for riddling. Dots and error bars represent the mean and 2.5-to-97.5 percentile range across $n=50$ $\varepsilon$-close pairs for each uncertainty $\varepsilon$. \textbf{b,} Same as (\textbf{a}), except at the sub-optimal learning rate of $\eta = 10^{-7}$. The churn is negligible (i.e., learned functions are essentially identical; $0.1\%$ on average), analogous to $f(\varepsilon) = 0$ for a non-riddled basin. The churn magnitude is the main distinction between inside and outside the riddling-like regime (Supplementary Sec.~6). \textbf{c,} Logarithmic derivatives of the trends in (\textbf{a}) and (\textbf{b}), showing the functional uncertainty exponent remains near-zero at small scales. \textbf{d,} Dots and error bars show the mean and full range of training loss across $n=5$ independent fine-tuning runs. \textbf{e,} Testing error. \textbf{f,} Loss convergence rate. \textbf{g,} Error convergence rate.}
    \label{fig:bert}
\end{figure}

\begin{figure}[h]
    \centering
    \includegraphics{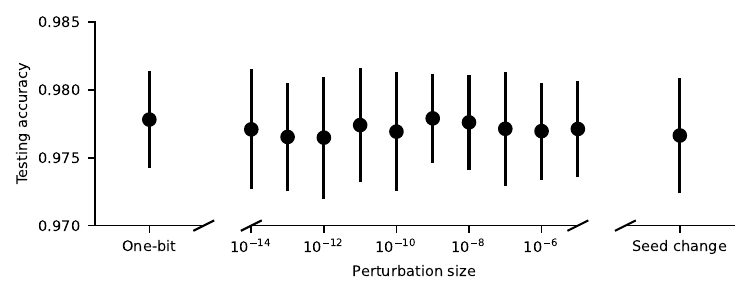}
    \caption{\textbf{Model variability is insensitive to perturbation size.} Dots and error bars show the mean and standard deviation of the testing accuracy across $n=100$ 30-epoch training runs with varying perturbations. The first point represents a flip to the least significant bit of a randomly selected parameter of the reference network in Fig.~\ref{fig:5}. The middle points apply random perturbations to each parameter of the reference network by a uniform random value $U(-\varepsilon, \varepsilon)$, where $\varepsilon$ is the perturbation size. The final point uses random Kaiming-initializations \cite{he2015delving} with different seeds.}
    \label{fig:variability}
\end{figure}

\begin{figure}[h]
    \centering
    \includegraphics{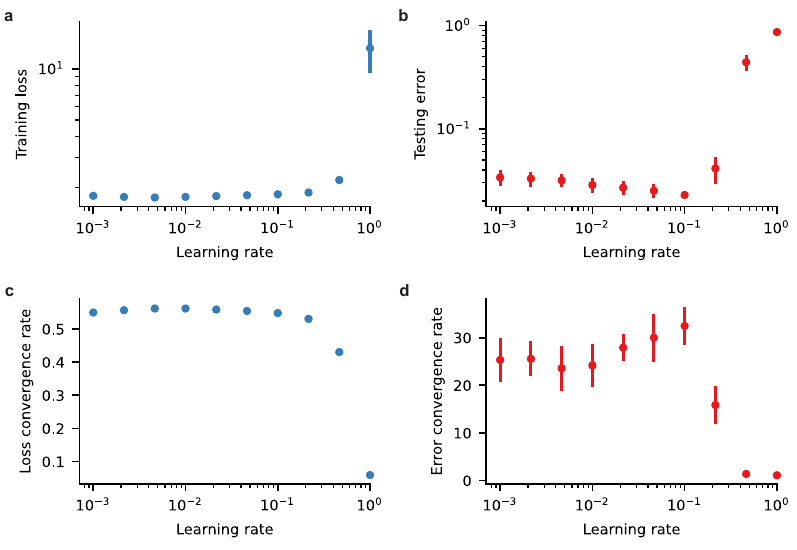}
    \caption{\textbf{Learning-rate sweep.} Dots and error bars show the mean and 95\% confidence interval across $n=5$ independent 100-epoch training runs on randomly initialized VGG-12 networks. \textbf{a,} Minimum training loss achieved during training. \textbf{b,} Minimum testing error achieved during training. \textbf{c,} Loss convergence rate defined as $\sum_t \bar{L}_t^{-1}$, where $\bar{L}_t$ denotes the average training loss in the $t$-th epoch. The larger the value, the longer the training spends with lower loss. \textbf{d,} Error convergence rate defined analogously using the testing error. Note that the generalization metrics in (\textbf{c}) and (\textbf{d}) are optimal at $\eta = 0.1$.}
    \label{fig:lr_sweep}
\end{figure}

\begin{figure}[h]
    \centering
    \includegraphics{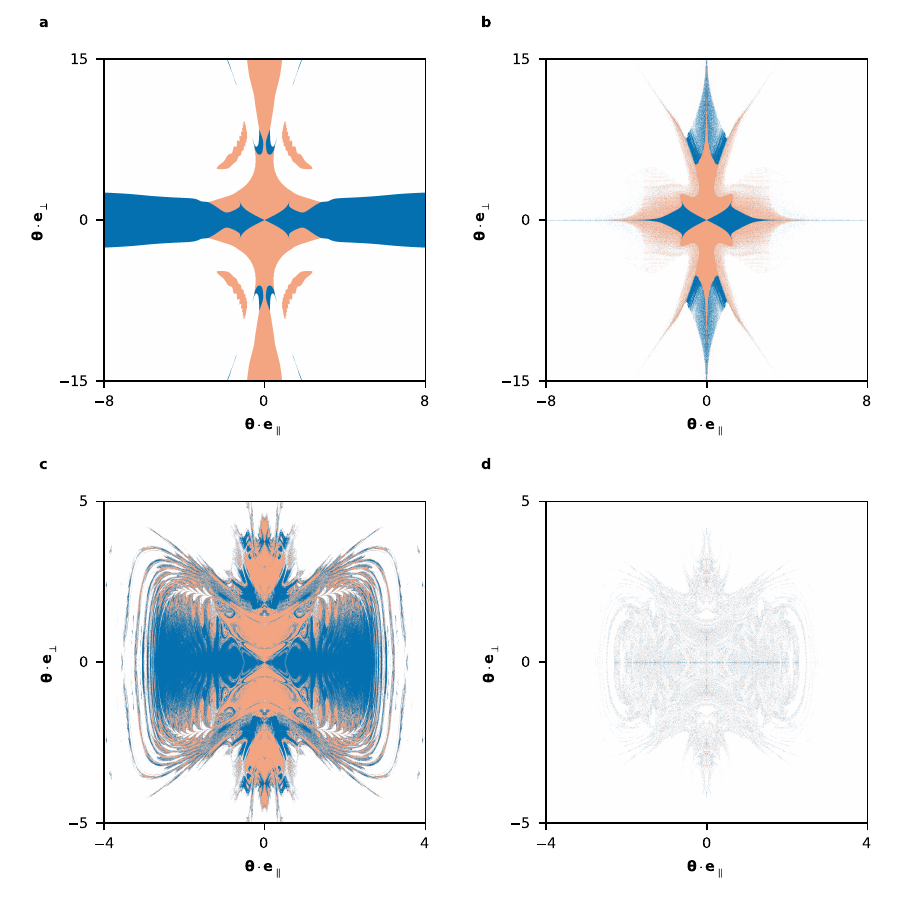}
    \caption{\textbf{Metamorphoses of riddling.} Destination maps for initializations in the same two-dimensional slice of parameter space as Fig.~\ref{fig:3}. Only the learning rate $\eta$ is varied; all other settings are held fixed. \textbf{a,} At $\eta = 0.1$, there is no riddling. \textbf{b,} At $\eta = 1$, there is pseudo-riddling, which is a mixture of open and riddled sets. \textbf{c,} At $\eta = 2.7$, there is true riddling. The structure is the same as Fig.~\ref{fig:3}\hyperref[fig:3]{b}. \textbf{d,} At $\eta = 3$, the time-dependent basin of a chaotic transient is riddled with diverging initializations. See Supplementary Sec.~1 and 2 for further information on these regimes.}
    \label{fig:robustness}
\end{figure}

\begin{figure}[h]
    \centering
    \includegraphics{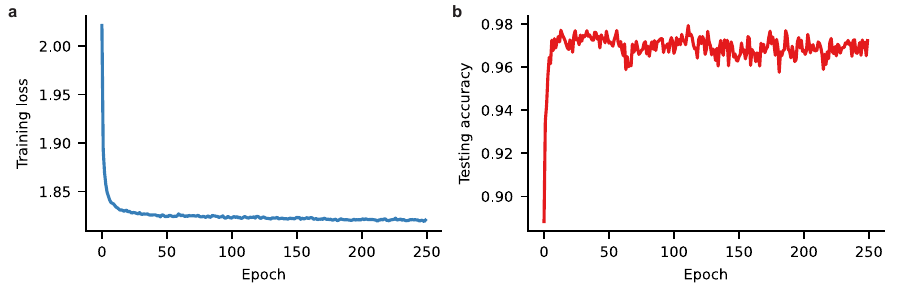}
    \caption{\textbf{Learning curves.} \textbf{a,} Loss on the training dataset at each epoch of training the randomly initialized VGG-12 network in Fig.~\ref{fig:4}\hyperref[fig:4]{a}. \textbf{b,} Same as (\textbf{a}), except with the accuracy on the testing dataset.}
    \label{fig:learning_curves}
\end{figure}

\end{document}